\documentclass[conference]{IEEEtran}
\IEEEoverridecommandlockouts
\usepackage{amsfonts}
\usepackage{dsfont}
\usepackage{mathtools, stmaryrd}
\usepackage{amssymb}
\usepackage{tikz}
\usepackage{tikzit}

\tikzstyle{blue dot}=[fill={rgb,255: red,127; green,193; blue,255}, draw=none, shape=circle, scale=0.3]
\tikzstyle{red dot}=[fill={rgb,255: red,255; green,127; blue,127}, draw=none, shape=circle, scale=0.3]
\tikzstyle{green dot}=[fill={rgb,255: red,127; green,201; blue,127}, draw=none, shape=circle, scale=0.3]
\tikzstyle{rectangle}=[fill=white, draw=black, shape=rectangle]
\tikzstyle{large box}=[fill=none, draw=black, shape=rectangle, minimum width=0.75cm, minimum height=1cm, rounded corners=2mm]
\tikzstyle{trapezium}=[fill={rgb,255: red,150; green,186; blue,218}, draw=none, shape=trapezium, minimum height=1cm, minimum width=1cm, rotate=270, rounded corners=0.2mm]
\tikzstyle{invisible large box}=[fill=none, draw=none, shape=rectangle, minimum width=0.75cm, minimum height=1cm]
\tikzstyle{dashed circle}=[fill=none, draw=black, shape=circle, minimum size=1.9cm, dashed]
\tikzstyle{star}=[fill=white, draw=black, shape=star, scale=0.3]

\tikzstyle{standard edge}=[draw=black, -, fill={rgb,255: red,215; green,217; blue,218}, fill opacity=0.2]
\tikzstyle{big dash}=[-, thick, dashed, dash pattern=on 4mm off 2mm, fill=none]
\tikzstyle{dashed edge}=[draw=black, -, dashed]
\tikzstyle{arrow}=[->, thick]
\tikzstyle{arrow2}=[{|->}]
\tikzstyle{arrow left}=[<-]
\tikzstyle{arrow left2}=[{<-|}]
\tikzstyle{arrow both sides}=[<->]
\tikzstyle{blue edge}=[-, draw=blue, thick]
\tikzstyle{blue arrow}=[draw=blue, ->, thick]
\tikzstyle{red edge}=[-, draw=red, thick]
\tikzstyle{dash points}=[-, dash pattern=on 0.01mm off 0.1mm, dashed]
\tikzstyle{thin edge}=[-, line width=0.01, fill={rgb,255: red,122; green,190; blue,232}, fill opacity=0.5, thick]
\tikzstyle{gray}=[-, draw={rgb,255: red,191; green,191; blue,191}]
\tikzstyle{edge}=[-, fill=white]
\tikzstyle{new edge style 0}=[-, fill=yellow, opacity=0.2, draw=none]
\tikzstyle{new edge style 1}=[-, fill={rgb,255: red,37; green,255; blue,255}, opacity=0.2, draw=none]
\tikzstyle{new edge style 2}=[-, fill={rgb,255: red,255; green,74; blue,137}, opacity=0.2, draw=none]

\usepackage[hyperindex,breaklinks]{hyperref}
\usepackage{bm}

\usepackage{paralist}
\usepackage{subfig}
\usepackage{pgfplots}
\usepackage{ftnxtra}
\usepackage{xparse} \DeclarePairedDelimiterX{\Iintv}[1]{\llbracket}{\rrbracket}{\iintvargs{#1}}
\NewDocumentCommand{\iintvargs}{>{\SplitArgument{1}{,}}m}
{\iintvargsaux#1} %
\NewDocumentCommand{\iintvargsaux}{mm} {#1\mkern1.5mu..\mkern1.5mu#2}
\DeclareMathOperator{\Tr}{Tr}

\usepackage{amsthm}

\newtheoremstyle{mystyle}
  {}
  {}
  {\itshape}
  {}
  {\bfseries}
  {.}
  { }
  {\thmname{#1}\thmnumber{ #2}\thmnote{ (#3)}}

\theoremstyle{mystyle}

\newtheorem{definition}{Definition}

\newtheorem{lemma}{Lemma}

\newcommand{\norm}[1]{\left\lVert#1\right\rVert}

\let\originalleft\left
\let\originalright\right
\renewcommand{\left}{\mathopen{}\mathclose\bgroup\originalleft}
\renewcommand{\right}{\aftergroup\egroup\originalright}
\newcommand\Mark[1]{\textsuperscript#1}

\usepackage{cite}
\usepackage{amsmath,amssymb,amsfonts}
\usepackage{algorithmic}
\usepackage{graphicx}
\usepackage{textcomp}
\usepackage{xcolor}
\def\BibTeX{{\rm B\kern-.05em{\sc i\kern-.025em b}\kern-.08em
    T\kern-.1667em\lower.7ex\hbox{E}\kern-.125emX}}
\begin{document}

\title{A Statistical Model for Predicting Generalization in Few-Shot Classification}

\author{\IEEEauthorblockN{Yassir Bendou\Mark{1}, Vincent Gripon\Mark{1}, Bastien Pasdeloup\Mark{1}, Giulia Lioi\Mark{1}, Lukas Mauch\Mark{2}, \\Stefan Uhlich\Mark{2}, Fabien Cardinaux\Mark{2}, Ghouthi Boukli Hacene\Mark{2}\Mark{3} and Javier Alonso Garcia\Mark{2}} \\
\IEEEauthorblockA{\Mark{1}IMT Atlantique, Lab-STICC, UMR CNRS 6285, F-29238 Brest, France} 
\IEEEauthorblockA{\Mark{2}Sony Europe, R\&D Center, Stuttgart Laboratory 1, Germany}
\IEEEauthorblockA{\Mark{3}Mila, Montréal, Canada}
\IEEEauthorblockA{\Mark{1}name.surname@imt-atlantique.fr, \Mark{2}name.surname@sony.com}}
\maketitle

\begin{abstract}
The estimation of the generalization error of classifiers often relies on a validation set. Such a set is hardly available in few-shot learning scenarios, a highly disregarded shortcoming in the field. In these scenarios, it is common to rely on features extracted from pre-trained neural networks combined with distance-based classifiers such as nearest class mean. In this work, we introduce a Gaussian model of the feature distribution. By estimating the parameters of this model, we are able to predict the generalization error on new classification tasks with few samples. We observe that accurate distance estimates between class-conditional densities are the key to accurate estimates of the generalization performance. Therefore, we propose an unbiased estimator for these distances and integrate it in our numerical analysis. We empirically show that our approach outperforms alternatives such as the leave-one-out cross-validation strategy. 
\end{abstract}

\begin{IEEEkeywords}
few-shot learning, classification, deep learning, generalization 
\end{IEEEkeywords}

\section{Introduction}
During the last decade, the problem of few-shot classification, that is to say classification with very few training samples (typically less than ten per class~\cite{miniimagenet}), has known a large number of contributions~\cite{laplacian, invariance_equivariance, squeezing, enhancing}. Many current state-of-the-art solutions consist in using a pre-trained deep feature extractor to embed samples in a feature space where classes are expected to be easier to discriminate. Then, distance-based classifiers such as nearest-class mean (NCM) are applied to the obtained feature vectors~\cite{easy, tian2020rethinking, simpleshot}. 
\begin{figure}[!h]
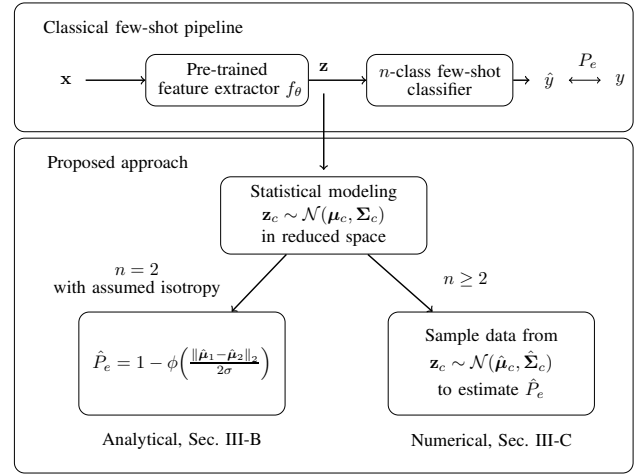

     \ctikzfig{tikz/pipeline}
    \caption{The classical few-shot pipeline consists in using a pre-trained feature extractor to embed data $\mathbf{x}$ into features $\mathbf{z}=f_\theta(\mathbf{x})$ and combine it with a classifier. Our approach consists of modeling the class-densities of the features as Gaussian distributions in a lower dimensional subspace and predicting the probability error of the classifier $\hat{P}_e$. Depending on the number of classes, we either use an analytical form of the probability error or estimate it by sampling a large number of datapoints.}  
    \label{fig:pipeline}
\end{figure}

In the few-shot scenario, performance prediction is not straightforward. The classical approach is to perform cross-validation which is hardly feasible with few training samples. Indeed, removing training samples and using them as a holdout test set severely impacts the generalization ability and at the same time, using too few such samples to measure accuracy on previously unseen data leads to poor performance estimates~\cite{biasedcrossval}. The usual strategy consists in ``leaving-one-out cross-validation''~\cite{crossvallimitations} where one averages the generalization estimated by removing a single training sample of each class over multiple such independent random removals. 

Finding an alternative to cross-validation to measure generalization ability has been extensively studied in the literature for standard classification settings where a large number of training samples is available~\cite{fantasticgeneralizationmeasures}. In this work, we are mainly interested in proposing an alternative to cross-validation and to existing generalization prediction methods in the context of few-shot classification. The proposed method estimates the parameters of a statistical model of the class-conditional densities at the output of the feature extractor. The model is then used to predict the probability of error (c.f. Figure~\ref{fig:pipeline}).

There are two key difficulties here: \begin{inparaenum}[1)]
    \item The first one is that we need to identify a statistical model that is both expressive enough to accurately capture the data distribution and at the same time depends on as few parameters as possible, such that we can accurately estimate them even in the very low data regime.
    \item The second one is that our derived expression for the probability of error depends on the distances between class centers. We observe that the naive estimate for the distances is biased, which leads to underestimating the probability of error especially when working in high-dimensional spaces with very few samples.
  \end{inparaenum}

The main contributions\footnote{The code to reproduce the results of our experiments is available in the following link: \url{https://github.com/ybendou/fs-generalization}.} of our work are as follows:

\begin{itemize}
\item We introduce a statistical model of class-conditional densities in the feature space;
\item Using this model, we obtain statistical bounds on the generalization error;
\item We show that the naive estimator for the distances between class centers is biased and, to alleviate this problem, we propose an unbiased estimator instead;
\item We provide experiments and show that our method outperforms other model-free generalization error predictors, such as the leaving-one-out cross-validation, in the context of standardized few-shot classification benchmarks.
\end{itemize}

\section{Related Work}
\label{section:relatedwork}
\subsection{Classification in few-shot learning}

Since there are few training samples in few-shot classification tasks, training a deep neural network architecture that is typically constructed from a large number of parameters is out of the question. The main approach is to use a pre-trained deep feature extractor and combine it with a simple classifier trained for the few-shot task. In this paper, we refer to $P$ for probabilities and $p$ for probability density functions. We formalize few-shot classification as follows:

\begin{definition}[Few-shot classification]

Let $\mathcal{D}_B~=~\{(\mathbf{x'}_i, {y'}_{i})\}_{i=1}^{m}$ be a large base dataset where $\forall i, (\mathbf{x'}_i, {y'}_i)$ are i.i.d samples drawn from the true joint probability distribution $p_{\mathcal{X}_B, \mathcal{Y}_B}$ and $\forall i,  {y'}_i~\in~\mathcal{Y}_B$ is the class associated with $\mathbf{x'}_i$. We are also given a small few-shot training dataset $\mathcal{D} = \{(\mathbf{x}_{j}, y_{j})\}_{j=1}^{\ell}$, where $\forall j, y_j\in \mathcal{Y} \neq \mathcal{Y}_B$ and $\forall j, (\mathbf{x}_{j}, y_{j})\sim p_{\mathcal{X},\mathcal{Y}}$.

The goal of few-shot classification is to train a classifier $C$ using $\mathcal{D}$, with the potential help of $\mathcal{D}_B$.
\end{definition}

In the literature, most methods consist in training a deep feature extractor $f_{\theta}$ with parameter set $\theta$ using $\mathcal{D}_B$. This feature extractor is then either adapted or used as is on $\mathcal{D}$ to produce feature vectors $\mathbf{z} = f_{\theta}(\mathbf{x})$ in a Euclidean space. The few-shot classifier then works with the modified training dataset $f_{\theta}(\mathcal{D}) = \{(\mathbf{z}_{j}, y_{j})\}_{j=1}^{\ell}$.

There are multiple strategies on how to train $f_\theta$ that can roughly be classified as optimization-based approaches and distance metric approaches.

Optimization-based methods aim to effectively adapt $f_\theta$ to new few-shot tasks. Meta-learning has been a popular method especially with the introduction of MAML and its variants~\cite{Finn2017, liu2020task,MetaDiffConv,fei2020melr,liu2020task} which aim to learn a good initialization of the neural network such that it can quickly adapt to new tasks with few gradient steps. On the other hand, distance-based approaches aim to learn a good feature extractor~\cite{dhillon2019baseline}. The main idea is to find an appropriate feature space where each image is well represented for transfer tasks.

In the distance-based category, the feature extractor can be trained in two different ways. The first one relies on episodic training where the idea is to reproduce the same conditions of the few-shot adaptation phase during the pre-training of the feature extractor~\cite{Snell2017, vinyals2016matching}. Episodic training is not specific to distance-based approaches as it is also used in meta-learning. The second way to train a feature extractor is to use a standard cross-entropy loss. This is usually referred to as transfer learning, which has been successful in recent years and largely adopted due to its competitive performance compared to episodic training, while being relatively simple to implement~\cite{simpleshot,cosoc,pal,easy}.

A common approach in transfer learning is to rely on data augmentation, self-supervision and manifold mixup~\cite{mangla2020charting} to learn a robust feature extractor for better generalization ability. In this paper, we adopt the transfer learning strategy with the above-mentioned training techniques.

Various few-shot classifiers have been proposed in the literature such as fine tuning a multi-layer perceptron with a cross-entropy loss~\cite{dhillon2019baseline}, which has been criticized for being biased in few-shot regimes~\cite{firthbias}. Other distance-based approaches such as using a nearest class mean classifier~\cite{simpleshot} or an earth distance metric using optimal transport~\cite{deepemd} have also been studied. We adhere to the nearest class mean (NCM) approach due to its well established performance and its simplicity. 

\begin{definition}[Nearest class mean classifier]
A nearest class mean classifier $C_{\text{NCM}}$ is the optimal classifier when class-conditional densities follow a Gaussian distribution with equal isotropic covariance and uniform prior across classes~\cite{ncmoptimal}:
\begin{equation}
    p(\mathbf{z}~|~y=c)=\mathcal{N}(\mathbf{z}~|~\boldsymbol{\mu}_c, \sigma^2\mathbf{I})
    \;,
\end{equation}
where $\boldsymbol{\mu}_c$ is the center of class $c \in \mathcal{Y}$, $\sigma$ is the standard deviation and $\mathbf{I}$ the identity matrix.
The classification of a new sample $\mathbf{z}$ is performed according to:
\begin{equation}
  C_{\text{NCM}}(\mathbf{z}) = \arg\min_{c \in \mathcal{Y}} \norm{\mathbf{z} - \boldsymbol{\mu}_c}_2
  \;.
  \label{eq:ncm}
\end{equation}
In practice we estimate the class centers from the training data $\mathcal{D}$ using the empirical average of each class. 
\end{definition}

Once the class centers are estimated, there are a maximum of $(n-1)$ dimensions of interest in the considered Euclidean space, which correspond to directions between class centers. Remaining ones can be disregarded, as they produce contributions that are orthogonal to the axes between class centers, and thus contribute equally to any distance computation in Equation~\eqref{eq:ncm}. Projecting the data onto this lower dimensional subspace is preferable in few-shot classification as it allows to work with lower dimensions~\cite{bavardage}. Such a projection can be performed using for example a QR decomposition~\cite{QR} to reduce the dimension of our data. Indeed, as shown in~\cite{bavardage}, a subspace of dimension $(n-1)$ does not impact the boundary decisions of a NCM classifier.

\subsection{Predicting Generalization}

Predicting generalization is one of the most important topics of ML. It was brought into focus by~\cite{randomlabels}, who asked the question of how one can measure the generalization from training data, showing that neural networks can easily fit randomly labeled data with high accuracy but with low generalization capabilities. We can define the problem of predicting generalization as follows~\cite{ntk}:

\begin{definition}[Predicting generalization]


Given an underlying joint probability distribution $p_{\mathcal{X},\mathcal{Y}}$ and a classifier $C$ trained on $\mathcal{D}$, the goal of predicting generalization is to estimate the error defined as: 
\begin{equation}
    \mathcal{R}(C) = \mathbb{E}_{(\mathbf{x},y)\sim p_{\mathcal{X},\mathcal{Y}}}[\mathds{1}_{C\left(\mathbf{x}\right)\neq y}]\;.
\end{equation}
\end{definition}


Many works have been proposed on generalization of a neural network trained on standard and large training datasets such as ImageNet~\cite{neuripschallenge} and on finding a measure which highly correlates with the accuracy instead of predicting the accuracy itself. In that regard, the Kendall's rank-correlation coefficient is commonly used~\cite{fantasticgeneralizationmeasures}, since usually the ordering between different models is what matters for applications such as finding the best hyper-parameters or the best architecture in the field of neural architecture search.

The proposed methods in the literature can be summarized into few different families. The first one is PAC-Bayes methods where the generalization behavior of a model is described by probably approximately correct (PAC) bounds~\cite{pac}; these methods often provide an upper-bound on the generalization error and the results are often restricted to a small set of models (\emph{e.g.}, no depth variations). The second family is norm-based methods, which analyze the neural network weights. These methods have shown to perform poorly~\cite{fantasticgeneralizationmeasures}. The last family of methods aims to analyze the intermediate representation of the training data in the feature space. Many methods have been proposed in this line of work such as using the Davies-Bouldin Index~\cite{dbindex} which is a clustering measure of the training data. Another approach~\cite{marginbounds} measures the distance of the training data to the decision boundaries in multiple intermediate features. Note that the main focus of these methods is to predict the generalization of a model trained for a certain task where large training data is available. Predicting generalization when working with few labeled samples has mainly been addressed when using meta-learning methods~\cite{farid2021generalization,chen2021generalization}. The closest work to ours is~\cite{myriam}, where some of the strategies for predicting generalization mentioned before have been tested for few-shot classification using transfer learning.

Differently to previously mentioned works, in this paper we aim at deriving a statistical model of the class-conditional densities in the feature space and to use this model to estimate the generalization error. As we will demonstrate in the experiments, the proposed methodology can outperform the previously mentioned ones in few-shot settings.

\section{Methodology}

\subsection{Statistical model}

As previously mentioned, the first step of our proposed methodology consists in proposing a statistical model for class-conditional densities in the feature space.

Let us assume that each class follows a Gaussian distribution with a uniform prior across the classes, i.e, $p(y_\mathbf{z}=c)=\frac{1}{n}, \forall c \in \mathcal{Y}$, where $n$ is the cardinal of $\mathcal{Y}$, which is reasonable to assume in a few-shot setting~\cite{bavardage}. The conditional densities are defined as: 
\begin{equation}
    p(\mathbf{z}~|~y=c)=\mathcal{N}(\mathbf{z}~|~\boldsymbol{\mu}_c, \boldsymbol{\Sigma}_c)
    \;,
\end{equation}
where $\boldsymbol{\Sigma}_c$ is the covariance matrix of class $c$.

Our hypothesis stands from the fact that given a well pre-trained feature extractor, each class in the feature space should follow a multivariate Gaussian distribution centered around a class center. This assumption has been largely adopted in the few-shot literature~\cite{gaussianfeatures,ptmap,gaussianfeaturesvae,cao2019theoretical}. Furthermore, the performance we obtain through our experimental results in Section~\ref{section:experiments} show that this model fits our data.  

Predicting generalization can be defined as predicting the probability of error from the classifier $C$. 
Let $R_c~=~\{\mathbf{z}~|~C(\mathbf{z})=c\}$ be the decision region for class $c$ using the classifier $C$ and $R~=~\cup_{c\in\mathcal{Y}} R_c$. The theoretical error of our problem is defined by the sum of integrals:
\begin{equation}
    P_e = \sum_{c \in\mathcal{Y}} \int_{R\smallsetminus R_c} p(\mathbf{z}~|~y=c)p(y=c)\,d\mathbf{z}\;. 
\label{eq:probaerror}
\end{equation} 

\subsection{Analytical insight}
\label{subsection:analytical}
A closed form solution of Equation~\ref{eq:probaerror} when $n~>~2$ is often intractable. Therefore, in this section we focus on the case of binary classification. We derive an analytical expression for $P_e$ and propose a statistical bound for its estimate.

For the case of a binary classifier of isotropic Gaussian data with equal covariance $\mathbf{\Sigma}_c = \sigma^2 \mathbf{I}_d, \forall c$, where $d$ is the dimension of $\mathbf{z}$, and class centers $\bm{\mu}_a$ and $\bm{\mu}_b$, the probability error has a closed form which only depends on the distance between the class centers $r~=~\norm{\bm{\mu}_a-\bm{\mu}_b}_2$ and the shared standard deviation: 
\begin{equation}
P_e = 1-\phi\left(\frac{r}{2\sigma}\right)\;,
\end{equation}
where $\phi$ is the cumulative distribution function of $\mathcal{N}(0, 1)$.

To estimate $P_e$, we typically estimate $r$ and $\sigma$ using i.i.d samples such that $\hat{r}=\norm{\hat{\bm{\mu}}_a-\hat{\bm{\mu}}_b}_2$, where $\hat{\bm{\mu}}$ is the empirical mean estimate.

\paragraph{Statistical bound of the probability error}Under this analytical form we derive in the univariate case a statistical bound for $\hat{P}_e~=~1~-~\phi\left(\frac{\hat{r}}{2\hat{\sigma}}\right)$.

\begin{lemma}
\label{lemma:lemma1}
Given two classes represented by two univariate isotropic distributions $\mathcal{N}_a(\mu_a, \sigma^2)$ and $\mathcal{N}_b(\mu_b, \sigma^2)$ with shared and known standard deviations, the true probability of error $P_e$ is bounded by: 
\begin{equation}
  P\left(\left|P_e-\hat{P_e}\right| \leq \frac{\phi^{'}(0)}{\sqrt{2k}}\left|\phi^{-1}\left(1-\frac{\alpha}{2}\right)\right|\right)\geq1-\alpha\;,  
\end{equation}

with a probability $1-\alpha$, where $\hat{P}_e$ is the probability error estimate from a sequence of $k$ i.i.d random variables from the two distributions with empirical mean estimates $\hat{\mu}_a$ and $\hat{\mu}_b$ , $\phi$ is the cumulative distribution function of $\mathcal{N}(0,1)$, $\phi^{-1}$ its inverse and $\phi^{'}$ its derivative, with $\phi^{'}(0)=\frac{1}{\sqrt{2\pi}}$.
\end{lemma}
The proof of Lemma~\ref{lemma:lemma1} can be found in the Appendix.
The bound for the true probability error is thus $\mathcal{O}(\frac{1}{\sqrt{k}})$. We compare this behavior on real data in the Appendix. Furthermore, this result holds for multivariate isotropic distributions as we can simply project on the axis between the two class centers to work with univariate distributions. In the remaining sections, we consider multivariate distributions.

\paragraph{Bias of the naive distance estimator}
Estimating the probability error depends on estimating the distance between class centers. The naive approach for estimating distances is usually performed by estimating the means of each distribution using the empirical mean estimate and computing $\hat{r}=\norm{\hat{\bm{\mu}}_a-\hat{\bm{\mu}}_b}_2$ to which we refer to as the naive estimator. However, this estimation of the distance between class centers is biased.

\begin{lemma}
\label{lemma:lemma2}
Let $(\bm{a}_1, \bm{a}_2, \cdots, \bm{a}_{k})$ and $(\bm{b}_1, \bm{b}_2, \cdots, \bm{b}_{k})$ be two sequences of i.i.d random variables drawn from their respective multivariate probability distributions $p_{a}$ and $p_{b}$ assumed independent with finite expected values $\bm{\mu}_a$ and $\bm{\mu}_b$ and finite second order moment with covariance matrices $\mathbf{\Sigma}_{a}$ and $\bm{\Sigma}_{b}$. Let $\hat{r}$ be the naive estimator for the distances using $\hat{\bm{\mu}}_{a}~=~\frac{1}{k} \sum_{i=1}^{k} \bm{a}_i$ and $\bm{\hat{\mu}}_{b}~=~\frac{1}{k} \sum_{i=1}^{k} \bm{b}_i$ the mean estimator of each of the two sequences, then:

\begin{equation}
    \mathbb{E}_{\substack{\bm{a}\sim p_a \\ \bm{b}\sim p_b}}\left(\hat{r}^2\right) - r^2 =  \frac{\Tr\left(\mathbf{\Sigma}_{a}+\mathbf{\Sigma}_{b}\right)}{k}\;.
\end{equation}

\end{lemma}

The proof of Lemma~\ref{lemma:lemma2} is included in the Appendix. This Lemma holds for any two independent distributions. 
The bias is a function of the noise and the number of samples $k$. 
In the case of two isotropic multivariate distributions $\mathcal{N}_{a}(\bm{\mu}_{a}, \sigma_{a}^2\mathbf{I}_d)$ and $\mathcal{N}_{b}(\bm{\mu}_{b}, \sigma_{b}^2\mathbf{I}_d)$, the bias becomes
\begin{equation}
  \mathbb{E}_{\substack{\bm{a}\sim \mathcal{N}_{a} \\ \bm{b}\sim \mathcal{N}_{b}}}\left(\hat{r}^2\right) - r^2 =  \frac{d\left(\sigma_{a}^2+\sigma_{b}^2\right)}{k}\;.  
\end{equation}

Interestingly, the bias scales with the ratio $\frac{d}{k}$. 
We include this bias reduction step as part of our numerical approach. The correction is performed in the original high dimensional space. We show the extent of this bias in the experimental results in section~\ref{section:experiments}. Furthermore, to show the importance of this step, in our experiments we provide results with and without correcting the bias.   

\subsection{Numerical insight}
\label{subsection:numerical}

Computing $P_e$ analytically in equation~\ref{eq:probaerror} for $n > 2$ is hard. However, in practice we can approximate $P_e$ through a Monte Carlo method. For each class, we draw a large number of data points from the Gaussian distributions that we fitted to the few-shot training dataset. This way, we artificially enrich our dataset and compute the classifier's decisions on a \emph{virtual validation set}. We sample in the reduced subspace with $n-1$ dimensions. Note that the decision regions in the lower dimensional and in the original subspace are the same, as already stated in Section~\ref{section:relatedwork}.

Similar to the binary case, we need to take into account that the distances between the estimated class centers are positively biased. This leads to underestimated $P_e$ (see Appendix). Correcting this bias is key to accurately estimating $P_e$. In fact, for a distance-based classifier, the absolute positioning of the class centers do not affect its decisions. In order to perform the sampling, we generate a set of points which respect the new estimated distances using the Nonmetric Multidimensional Scaling algorithm~\cite{MDS}.

Estimating $P_e$ by sampling means that there are no restrictions for choosing the covariance of the data. In section~\ref{section:experiments} we compare the performance for different covariance matrices, i.e.: \begin{inparaenum}[1)]
    \item using the identity matrix,
    \item using a shared isotropic covariance matrix across classes,
    \item using isotropic covariance matrix per class,
    \item using full covariance matrix per class.
\end{inparaenum} In section~\ref{section:experiments} we run an experiment to validate our choice.

\section{Experiments}
\label{section:experiments}
\subsection{Datasets and implementation details}
We run our experiments on three standardized few-shot vision classification benchmarks:

\begin{itemize}
\item Mini-ImageNet~\cite{miniimagenet}: A dataset for in-domain few-shot learning extracted from ImageNet. The dataset is divided into three splits with disjoint classes. The first split with 64 classes (base classes) is used to train the feature extractor, the second split contains 16 classes for validation and the third split contains 20 classes (novel classes) for test. Each class contains 600 images.

\item Tiered-ImageNet~\cite{tieredimagenet}: Another subset of ImageNet for in-domain classification with 351 base classes, 97 classes for validation and 160 novel classes. Each class contains around 1300 samples. This dataset has classes with a hierarchy structure.

\item Meta-dataset~\cite{metadatasets}: The classical Meta-dataset benchmark consists of 10 different datasets from which we use 4 for our experiments. The first dataset is ImageNet for in-domain classification which is also split into three disjoint subsets for training the feature extractor, validating and testing. The remaining 3 datasets are for cross-domain classification (VGG Flower, CUB-200-2011 and Describable Textures). Note that in cross-domain the training split of ImageNet is used to train the feature extractor and the 3 remaining datasets are used for testing. 
\end{itemize}

Each of these benchmarks contains a large number of samples ($\geq$ 500 samples per class) from which we artificially and randomly sample, for each dataset, $10^3$ few-shot classification problems to run our experiments.

We use the standard training procedure proposed in~\cite{mangla2020charting} to pre-train a ResNet-18 architecture~\cite{resnet}. This architecture along the training procedure has been vastly used in the few-shot classification literature with feature vectors of 512 dimensions~\cite{easy,dhillon2019baseline}.

\subsection{Estimating the first and second order moments}
\label{subsection:Frobenius}
\begin{figure}[!htbp]
    \centering
    \scalebox{1}{
\begin{tikzpicture}

\definecolor{dodgerblue0143213}{RGB}{0,143,213}
\definecolor{goldenrod22917456}{RGB}{229,174,56}
\definecolor{lightgray203}{RGB}{203,203,203}
\definecolor{lightgray204}{RGB}{204,204,204}
\definecolor{olivedrab10914479}{RGB}{109,144,79}
\definecolor{tomato2527948}{RGB}{252,79,48}
\definecolor{whitesmoke240}{RGB}{240,240,240}

\begin{axis}[
axis background/.style={fill=white},
axis line style={whitesmoke240},
legend cell align={left},
legend style={fill opacity=0.8, draw opacity=1, text opacity=1, draw=lightgray204, fill=white, at={(0.97,1.2)}},
tick align=outside,
tick pos=left,
title={},
unbounded coords=jump,
x grid style={lightgray203},
xlabel={Number of samples per class},
xmajorgrids,
xmin=-1.45, xmax=52.45,
xtick style={color=black},
y grid style={lightgray203},
ylabel={KL divergence},
ymajorgrids,
ymin=-2, ymax=25,
ytick style={color=black},
yscale=0.8
]
\addplot [ultra thick, dodgerblue0143213, mark size=1.5pt, mark=*]
table {%
1 16.1103420257568
2 11.2338943481445
3 9.62686347961426
4 8.85332489013672
5 8.29176712036133
6 7.97358465194702
7 7.80102348327637
8 7.61294507980347
9 7.45868539810181
10 7.32281494140625
11 7.24482917785645
12 7.15517997741699
13 7.09883260726929
14 7.02899169921875
15 6.97849559783936
16 6.93380308151245
17 6.90262079238892
18 6.85727119445801
19 6.82847213745117
20 6.80037069320679
21 6.77842044830322
22 6.76115798950195
23 6.7373948097229
24 6.71790838241577
25 6.70621776580811
26 6.68804550170898
27 6.66571950912476
28 6.65676021575928
29 6.64397048950195
30 6.63187074661255
31 6.61785459518433
32 6.60750722885132
33 6.59519529342651
34 6.58869886398315
35 6.58502864837646
36 6.57542181015015
37 6.57064008712769
38 6.56028366088867
39 6.55240440368652
40 6.55117177963257
41 6.54539728164673
42 6.53725099563599
43 6.5299129486084
44 6.52223491668701
45 6.51495361328125
46 6.51042890548706
47 6.50665092468262
48 6.50366401672363
49 6.49904680252075
50 6.49530935287476
};
\addlegendentry{$(1)\hat{\mathbf{\Sigma}}=\mathbf{I}$}
\addplot [ultra thick, tomato2527948, mark size=1.5pt, mark=diamond*]
table {%
1 inf
2 7.93042850494385
3 5.70705556869507
4 4.95476150512695
5 4.43214416503906
6 4.10984516143799
7 3.98998403549194
8 3.8343780040741
9 3.78178954124451
10 3.65712213516235
11 3.57912850379944
12 3.49020671844482
13 3.48164296150208
14 3.42230701446533
15 3.37389349937439
16 3.33973073959351
17 3.33998203277588
18 3.29666233062744
19 3.29206538200378
20 3.27129578590393
21 3.25812101364136
22 3.23823070526123
23 3.21235680580139
24 3.20476078987122
25 3.19870066642761
26 3.18686890602112
27 3.18279409408569
28 3.16946220397949
29 3.15664315223694
30 3.13473749160767
31 3.12557983398438
32 3.12671113014221
33 3.10616493225098
34 3.09948182106018
35 3.09968852996826
36 3.11423587799072
37 3.10449504852295
38 3.09913873672485
39 3.09208679199219
40 3.08929347991943
41 3.0922863483429
42 3.07353615760803
43 3.0693154335022
44 3.06088590621948
45 3.06275320053101
46 3.06274652481079
47 3.05972003936768
48 3.05393362045288
49 3.05017375946045
50 3.04366731643677
};
\addlegendentry{$(2)\hat{\mathbf{\Sigma}}=\hat{\sigma}^2\mathbf{I}$}
\addplot [ultra thick, goldenrod22917456, mark size=1.5pt, mark=square*]
table {%
1 nan
2 11.2023067474365
3 7.56110095977783
4 6.33148765563965
5 5.50263977050781
6 4.99801445007324
7 4.70031499862671
8 4.4263014793396
9 4.2896523475647
10 4.10169076919556
11 3.97094082832336
12 3.82427906990051
13 3.77803564071655
14 3.68828940391541
15 3.61043548583984
16 3.55119681358337
17 3.53654479980469
18 3.47738099098206
19 3.45264196395874
20 3.42597270011902
21 3.39977955818176
22 3.38064646720886
23 3.34889531135559
24 3.33034038543701
25 3.30386328697205
26 3.27711462974548
27 3.26642513275146
28 3.24082612991333
29 3.21751499176025
30 3.19149851799011
31 3.17764759063721
32 3.17167901992798
33 3.14600849151611
34 3.13409304618835
35 3.12573575973511
36 3.13962960243225
37 3.12207245826721
38 3.11638236045837
39 3.10299968719482
40 3.09573483467102
41 3.09459137916565
42 3.07073998451233
43 3.0645318031311
44 3.05404424667358
45 3.05401492118835
46 3.04805326461792
47 3.03739523887634
48 3.03078007698059
49 3.02539801597595
50 3.01912117004395
};
\addlegendentry{$(3)\hat{\mathbf{\Sigma}}=\hat{\sigma}_c^2\mathbf{I}$}
\addplot [ultra thick, olivedrab10914479, mark size=1.5pt, mark=triangle*]
table {%
1 nan
2 nan
3 nan
4 33.0178070068359
5 14.5743637084961
6 9.70769596099854
7 7.59937810897827
8 6.38462781906128
9 5.46900796890259
10 4.79003095626831
11 4.25990343093872
12 3.89110565185547
13 3.56565499305725
14 3.26902770996094
15 3.01637578010559
16 2.80612540245056
17 2.65370512008667
18 2.48769307136536
19 2.34151411056519
20 2.24043440818787
21 2.1308867931366
22 2.02861571311951
23 1.93332469463348
24 1.85900926589966
25 1.7838978767395
26 1.69849276542664
27 1.65122449398041
28 1.5848902463913
29 1.53253519535065
30 1.48793840408325
31 1.43440747261047
32 1.39298188686371
33 1.34555208683014
34 1.30909776687622
35 1.27907955646515
36 1.25032866001129
37 1.22850072383881
38 1.19432985782623
39 1.16400337219238
40 1.14806294441223
41 1.12499701976776
42 1.09500050544739
43 1.06669306755066
44 1.0427680015564
45 1.0224232673645
46 1.00307476520538
47 0.98577606678009
48 0.967447102069855
49 0.949644148349762
50 0.934365749359131
};
\addlegendentry{$(4)\hat{\mathbf{\Sigma}}=\hat{\mathbf{\Sigma}}_c$}
\end{axis}

\end{tikzpicture}}
    \caption{Average KL divergence between a Gaussian distribution fitted with a limited number of samples and the closest Gaussian approximation using a larger number of samples. We average over $10^3$ 5-class few-shot classification problems from the test set of Meta-dataset ImageNet.}
    \label{fig:resultsKL}
\end{figure}
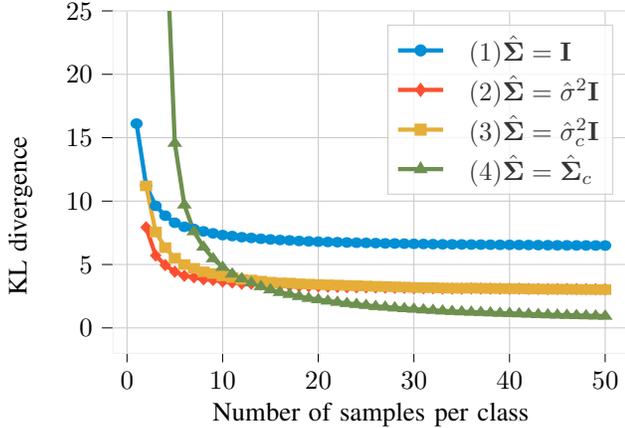

First, we conduct an experiment to validate the choice of the statistical model: particularly, which covariance matrix should be selected for the Gaussian model. We generate $10^3$ few-shot problems. For each of these, we estimate the mean and covariance matrix of each class under different models and compare it with the true distribution of each class from all the available samples in the original dataset (typically $\geq 500$ per class). We use the Kullback-Leibler (KL) divergence as a metric and vary the number of samples per class. The evolution of the error per model is given in Figure~\ref{fig:resultsKL}.

Figure~\ref{fig:resultsKL} shows a trade-off between the expressive power of each model and its capacity to overfit. As the number of samples per class increases ($\geq14$), the error of the model with more parameters (model 4) becomes lower than simpler models and hence it becomes a better choice. For very few samples, it is desirable to choose a simpler model with fewer parameters (model 2). Therefore, in our experiments we vary the model used depending on the number of samples. We use a shared isotropic covariance matrix across classes (model 2) for $k\leq(n-1)^2$ which is 16 for 5 class classification tasks and is roughly where the intersection between the model 2 and 4 is in Figure~\ref{fig:resultsKL}. For more samples, we use a free covariance matrix (model 4). 

\subsection{Unbiased estimator for the distances}
\begin{figure}[!htbp]
    \centering
   \scalebox{1}{\input{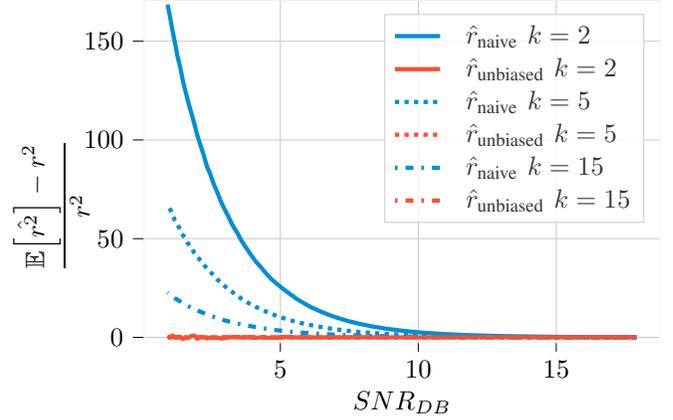}}
    \caption{Bias of the naive estimator. Here, the task is to estimate the distance between the center of two normal distributions. The true generative parameters are denoted $r$ and $\sigma$. We normalize the bias by the true distance and plot it against the SNR for different sample sizes $k$. The proposed estimator is unbiased while the naive estimator has a large bias especially in low SNR regimes with low samples.}
    \label{fig:resultsBias}
\end{figure}

In order to quantify the bias of the naive estimator for the distances from Lemma~\ref{lemma:lemma2}, we generate synthetic data from two isotropic distributions $\mathcal{N}\left(\bm{\mu}_1, \sigma^2\mathbf{I}_{d}\right)$ and $\mathcal{N}\left(\bm{\mu}_2, \sigma^2\mathbf{I}_d\right)$ under different settings of Signal-to-Noise ratios $\text{SNR}_{\text{DB}}~=~10\log_{10}\left(\frac{r}{\sqrt{2}\sigma}\right)$, where $r~=~\norm{\bm{\mu}_1-\bm{\mu}_2}_2$. Figure~\ref{fig:resultsBias} shows the bias of the naive estimator for the distance and the proposed unbiased estimator $\hat{r}^2_{\text{unbiased}}~=~\hat{r}^2_{\text{naive}}-\frac{2d\hat{\sigma}^2}{k}$, where $k$ is the number of available samples and $d=512$, the number of dimensions of the original features in our data. For a low SNR, the bias of the naive estimator becomes very important. On the other hand, the proposed estimator is unbiased. When estimating means from few data points, the estimated distance can be seen as the sum of the distance between the true means and the added distance in the orthogonal directions to the axis between the true means due to noise.
\subsection{Performance in predicting generalization}

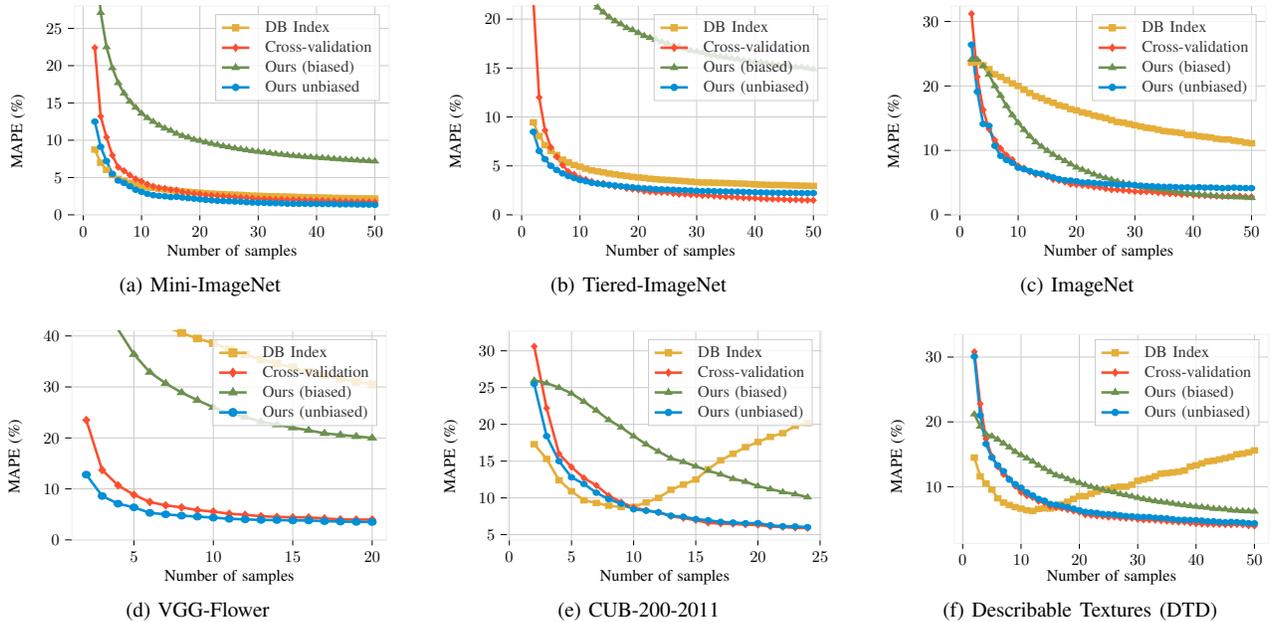
\begin{figure*}[!t]
\centering
\subfloat[Mini-ImageNet]{\scalebox{0.61}{{
\begin{tikzpicture}

\definecolor{dodgerblue0143213}{RGB}{0,143,213}
\definecolor{goldenrod22917456}{RGB}{229,174,56}
\definecolor{lightgray203}{RGB}{203,203,203}
\definecolor{tomato2527948}{RGB}{252,79,48}
\definecolor{whitesmoke240}{RGB}{240,240,240}
\definecolor{olivedrab10914479}{RGB}{109,144,79}

\begin{axis}[
axis background/.style={fill=white},
axis line style={whitesmoke240},
tick align=outside,
tick pos=left,
title={},
legend cell align={left},
legend style={fill opacity=0.8, draw opacity=1, text opacity=1, draw=lightgray203, fill=white, at={(0.97,1.2)}},
x grid style={lightgray203},
xlabel={Number of samples},
xmajorgrids,
xmin=-1.45, xmax=52.45,
xtick style={color=black},
y grid style={lightgray203},
ylabel={MAPE (\%)},
ymajorgrids,
ymin=0, ymax=28,
ytick style={color=black},
yscale=0.8
]
\addplot [ultra thick, goldenrod22917456, mark size=1.5pt, mark=square*]
table {%
2 8.74
3 6.98
4 6.02
5 5.33
6 4.82
7 4.54
8 4.29
9 4.08
10 3.89
11 3.73
12 3.63
13 3.5
14 3.4
15 3.29
16 3.22
17 3.14
18 3.09
19 3.02
20 2.97
21 2.89
22 2.84
23 2.8
24 2.77
25 2.73
26 2.7
27 2.67
28 2.62
29 2.57
30 2.55
31 2.52
32 2.49
33 2.48
34 2.46
35 2.45
36 2.42
37 2.39
38 2.37
39 2.35
40 2.35
41 2.33
42 2.3
43 2.27
44 2.26
45 2.24
46 2.24
47 2.22
48 2.21
49 2.2
50 2.18
};
\addlegendentry{DB Index}
\addplot [ultra thick, tomato2527948, mark size=1.5pt, mark=diamond*]
table {%
2 22.4
3 13.2
4 10.4
5 7.98
6 6.39
7 5.9
8 5.31
9 4.81
10 4.5
11 4.1
12 3.81
13 3.62
14 3.53
15 3.37
16 3.31
17 3.08
18 2.99
19 2.86
20 2.81
21 2.66
22 2.62
23 2.57
24 2.5
25 2.44
26 2.38
27 2.32
28 2.27
29 2.23
30 2.13
31 2.11
32 2.09
33 2.05
34 2.05
35 2.01
36 2.02
37 1.99
38 1.96
39 1.95
40 1.93
41 1.91
42 1.86
43 1.86
44 1.84
45 1.8
46 1.76
47 1.75
48 1.75
49 1.72
50 1.71
};
\addlegendentry{Cross-validation}

\addplot [ultra thick, olivedrab10914479, mark size=1.5pt, mark=triangle]
table {%
2 37
3 27.1
4 22.5
5 19.7
6 17.7
7 16.3
8 15.2
9 14.4
10 13.6
11 13
12 12.5
13 12
14 11.6
15 11.3
16 10.9
17 10.6
18 10.4
19 10.1
20 9.95
21 9.73
22 9.55
23 9.4
24 9.23
25 9.09
26 8.94
27 8.82
28 8.69
29 8.58
30 8.48
31 8.39
32 8.3
33 8.19
34 8.1
35 8.02
36 7.94
37 7.88
38 7.8
39 7.75
40 7.69
41 7.64
42 7.59
43 7.52
44 7.47
45 7.41
46 7.36
47 7.32
48 7.27
49 7.23
50 7.19
};
\addlegendentry{Ours (biased)}

\addplot [ultra thick, dodgerblue0143213, mark size=1.5pt, mark size=1.5pt, mark=*]
table {%
2 12.5
3 9.12
4 7.2
5 5.48
6 4.6
7 4.28
8 3.83
9 3.36
10 3.08
11 2.82
12 2.64
13 2.54
14 2.49
15 2.38
16 2.42
17 2.32
18 2.24
19 2.17
20 2.09
21 2.01
22 1.95
23 1.88
24 1.85
25 1.82
26 1.78
27 1.76
28 1.69
29 1.65
30 1.62
31 1.61
32 1.57
33 1.56
34 1.54
35 1.49
36 1.48
37 1.49
38 1.48
39 1.46
40 1.47
41 1.45
42 1.45
43 1.44
44 1.4
45 1.41
46 1.41
47 1.38
48 1.38
49 1.36
50 1.34
};
\addlegendentry{Ours unbiased}
\end{axis}
\end{tikzpicture}}}}\hfil 
\subfloat[Tiered-ImageNet]{\scalebox{0.61}{{
\begin{tikzpicture}

\definecolor{dodgerblue0143213}{RGB}{0,143,213}
\definecolor{goldenrod22917456}{RGB}{229,174,56}
\definecolor{lightgray203}{RGB}{203,203,203}
\definecolor{lightgray204}{RGB}{204,204,204}
\definecolor{tomato2527948}{RGB}{252,79,48}
\definecolor{whitesmoke240}{RGB}{240,240,240}
\definecolor{olivedrab10914479}{RGB}{109,144,79}

\begin{axis}[
axis background/.style={fill=white},
axis line style={whitesmoke240},
legend cell align={left},
legend style={fill opacity=0.8, draw opacity=1, text opacity=1, draw=lightgray204, fill=white, at={(0.97,1.2)}},
tick align=outside,
tick pos=left,
x grid style={lightgray203},
xlabel={Number of samples},
xmajorgrids,
xmin=-1.45, xmax=52.45,
xtick style={color=black},
y grid style={lightgray203},
ylabel={MAPE (\%)},
ymajorgrids,
ymin=0, ymax=21.3405,
ytick style={color=black},
yscale=0.8
]
\addplot [ultra thick, goldenrod22917456, mark size=1.5pt, mark=square*]
table {%
2 9.45
3 8.05
4 7.12
5 6.53
6 6.11
7 5.64
8 5.37
9 5.09
10 4.92
11 4.71
12 4.54
13 4.44
14 4.34
15 4.22
16 4.13
17 4.04
18 3.97
19 3.89
20 3.82
21 3.77
22 3.69
23 3.64
24 3.59
25 3.56
26 3.52
27 3.48
28 3.44
29 3.39
30 3.34
31 3.3
32 3.3
33 3.28
34 3.26
35 3.25
36 3.22
37 3.2
38 3.18
39 3.15
40 3.12
41 3.09
42 3.07
43 3.05
44 3.05
45 3.04
46 3.02
47 3
48 2.98
49 2.97
50 2.95
};
\addlegendentry{DB Index}
\addplot [ultra thick, tomato2527948, mark size=1.5pt, mark=diamond*]
table {%
2 21.9
3 12
4 8.63
5 6.89
6 5.94
7 5.09
8 4.41
9 4.13
10 3.76
11 3.61
12 3.4
13 3.23
14 3.22
15 3.02
16 2.9
17 2.9
18 2.7
19 2.63
20 2.56
21 2.43
22 2.37
23 2.32
24 2.3
25 2.28
26 2.19
27 2.11
28 2.12
29 2.06
30 2.05
31 1.98
32 1.98
33 1.93
34 1.89
35 1.84
36 1.83
37 1.79
38 1.74
39 1.72
40 1.7
41 1.65
42 1.63
43 1.61
44 1.6
45 1.57
46 1.53
47 1.54
48 1.51
49 1.5
50 1.48
};
\addlegendentry{Cross-validation}

\addplot [ultra thick, olivedrab10914479, mark size=1.5pt, mark=triangle]
table {%
2 53.1
3 40.4
4 34.4
5 30.8
6 28.3
7 26.6
8 25.2
9 24.1
10 23.2
11 22.4
12 21.8
13 21.2
14 20.7
15 20.2
16 19.8
17 19.5
18 19.1
19 18.9
20 18.6
21 18.3
22 18.1
23 17.9
24 17.7
25 17.5
26 17.3
27 17.2
28 17
29 16.8
30 16.7
31 16.6
32 16.4
33 16.3
34 16.2
35 16.1
36 16
37 15.9
38 15.8
39 15.7
40 15.6
41 15.6
42 15.5
43 15.4
44 15.3
45 15.3
46 15.2
47 15.1
48 15.1
49 15
50 14.9
};
\addlegendentry{Ours (biased)}

\addplot [ultra thick, dodgerblue0143213, mark size=1.5pt, mark size=1.5pt, mark=*]
table {%
2 8.48
3 6.52
4 5.69
5 5.01
6 4.58
7 4.23
8 3.96
9 3.75
10 3.55
11 3.43
12 3.27
13 3.19
14 3.12
15 3.06
16 3
17 2.92
18 2.87
19 2.81
20 2.75
21 2.71
22 2.68
23 2.62
24 2.6
25 2.58
26 2.56
27 2.53
28 2.51
29 2.49
30 2.45
31 2.42
32 2.43
33 2.41
34 2.39
35 2.38
36 2.37
37 2.36
38 2.34
39 2.32
40 2.3
41 2.28
42 2.27
43 2.26
44 2.25
45 2.25
46 2.24
47 2.23
48 2.22
49 2.22
50 2.22
};
\addlegendentry{Ours (unbiased)}
\end{axis}

\end{tikzpicture}}}}\hfil 
\subfloat[ImageNet]{\scalebox{0.61}{{
\begin{tikzpicture}

\definecolor{dodgerblue0143213}{RGB}{0,143,213}
\definecolor{goldenrod22917456}{RGB}{229,174,56}
\definecolor{lightgray203}{RGB}{203,203,203}
\definecolor{lightgray204}{RGB}{204,204,204}
\definecolor{tomato2527948}{RGB}{252,79,48}
\definecolor{whitesmoke240}{RGB}{240,240,240}
\definecolor{olivedrab10914479}{RGB}{109,144,79}

\begin{axis}[
axis background/.style={fill=white},
axis line style={whitesmoke240},
legend cell align={left},
legend style={fill opacity=0.8, draw opacity=1, text opacity=1, draw=lightgray204, fill=white, at={(0.97,1.2)}},
tick align=outside,
tick pos=left,
x grid style={lightgray203},
xlabel={Number of samples},
xmajorgrids,
xmin=-1.45, xmax=52.45,
xtick style={color=black},
y grid style={lightgray203},
ylabel={MAPE (\%)},
ymajorgrids,
ymin=0, ymax=32.4045,
ytick style={color=black},
yscale=0.8
]
\addplot [ultra thick, goldenrod22917456, mark size=1.5pt, mark=square*]
table {%
2 23.6
3 23.7
4 23.2
5 22.6
6 21.8
7 21.4
8 20.9
9 20.4
10 20
11 19.4
12 18.9
13 18.4
14 18.1
15 17.7
16 17.4
17 17
18 16.8
19 16.4
20 16.2
21 15.9
22 15.7
23 15.4
24 15.2
25 15
26 14.7
27 14.4
28 14.3
29 14.1
30 13.9
31 13.7
32 13.5
33 13.4
34 13.2
35 13
36 12.9
37 12.8
38 12.7
39 12.4
40 12.4
41 12.2
42 12.1
43 12
44 11.8
45 11.7
46 11.7
47 11.5
48 11.4
49 11.2
50 11.1
};
\addlegendentry{DB Index}

\addplot [ultra thick, tomato2527948, mark size=1.5pt, mark=diamond*]
table {%
2 31.2
3 21.4
4 16.3
5 13.3
6 11.6
7 10.3
8 9.25
9 8.57
10 7.58
11 7.25
12 6.66
13 6.27
14 6.22
15 5.9
16 5.48
17 5.32
18 5.09
19 4.76
20 4.69
21 4.6
22 4.5
23 4.32
24 4.26
25 4.07
26 3.9
27 3.91
28 3.81
29 3.74
30 3.63
31 3.57
32 3.59
33 3.56
34 3.44
35 3.38
36 3.28
37 3.27
38 3.16
39 3.18
40 3.09
41 3.03
42 2.99
43 2.96
44 2.91
45 2.87
46 2.89
47 2.9
48 2.91
49 2.8
50 2.79
};
\addlegendentry{Cross-validation}
\addplot [ultra thick, olivedrab10914479, mark=triangle*, mark size=1.5pt]
table {%
2 24.1
3 24.1
4 23.1
5 21.8
6 20
7 18.5
8 16.9
9 15.5
10 14.3
11 13.2
12 12.2
13 11.3
14 10.6
15 9.95
16 9.35
17 8.86
18 8.38
19 7.87
20 7.37
21 7.02
22 6.65
23 6.29
24 5.98
25 5.71
26 5.48
27 5.19
28 4.95
29 4.75
30 4.58
31 4.39
32 4.24
33 4.08
34 3.99
35 3.87
36 3.75
37 3.63
38 3.49
39 3.38
40 3.26
41 3.18
42 3.09
43 3.03
44 2.95
45 2.89
46 2.82
47 2.78
48 2.76
49 2.72
50 2.64
};
\addlegendentry{Ours (biased)}
\addplot [ultra thick, dodgerblue0143213, mark size=1.5pt, mark=*]
table {%
2 26.4
3 19.1
4 14.1
5 13.8
6 10.7
7 9.15
8 8.51
9 8.07
10 7.31
11 7.07
12 6.8
13 6.5
14 6.37
15 6.07
16 5.81
17 5.55
18 5.37
19 5.32
20 5.21
21 5.02
22 5.03
23 4.95
24 4.86
25 4.78
26 4.78
27 4.66
28 4.62
29 4.69
30 4.61
31 4.53
32 4.47
33 4.44
34 4.39
35 4.34
36 4.35
37 4.3
38 4.25
39 4.26
40 4.23
41 4.29
42 4.22
43 4.24
44 4.2
45 4.13
46 4.17
47 4.22
48 4.17
49 4.14
50 4.14
};
\addlegendentry{Ours (unbiased)}
\end{axis}

\end{tikzpicture}}}} 

\subfloat[VGG-Flower]{\scalebox{0.61}{{
\begin{tikzpicture}

\definecolor{dodgerblue0143213}{RGB}{0,143,213}
\definecolor{goldenrod22917456}{RGB}{229,174,56}
\definecolor{lightgray203}{RGB}{203,203,203}
\definecolor{lightgray204}{RGB}{204,204,204}
\definecolor{tomato2527948}{RGB}{252,79,48}
\definecolor{whitesmoke240}{RGB}{240,240,240}
\definecolor{olivedrab10914479}{RGB}{109,144,79}

\begin{axis}[
axis background/.style={fill=white},
axis line style={whitesmoke240},
legend cell align={left},
legend style={fill opacity=0.8, draw opacity=1, text opacity=1, draw=lightgray204, fill=white, at={(0.97,1.2)}},
tick align=outside,
tick pos=left,
x grid style={lightgray203},
xlabel={Number of samples},
xmajorgrids,
xmin=1.1, xmax=20.9,
xtick style={color=black},
y grid style={lightgray203},
ylabel={MAPE (\%)},
ymajorgrids,
ymin=0, ymax=40.966,
ytick style={color=black},
yscale=0.8
]
\addplot [ultra thick, goldenrod22917456, mark=square*]
table {%
2 50.7
3 48.8
4 46.7
5 44.9
6 43.1
7 41.9
8 40.6
9 39.5
10 38.6
11 37.4
12 36.4
13 35.4
14 34.6
15 33.8
16 33
17 32.3
18 31.6
19 31
20 30.5
};
\addlegendentry{DB Index}

\addplot [ultra thick, tomato2527948, mark=diamond*]
table {%
2 23.5
3 13.7
4 10.7
5 8.85
6 7.44
7 6.77
8 6.38
9 5.83
10 5.55
11 5.13
12 4.92
13 4.64
14 4.51
15 4.42
16 4.4
17 4.21
18 4.02
19 3.98
20 3.98
};
\addlegendentry{Cross-validation}

\addplot [ultra thick, olivedrab10914479, mark=triangle]
table {%
2 63.8
3 49.2
4 41.3
5 36.4
6 32.9
7 30.7
8 28.9
9 27.4
10 26
11 24.9
12 24.1
13 23.2
14 22.6
15 22
16 21.5
17 20.9
18 20.6
19 20.3
20 20
};
\addlegendentry{Ours (biased)}
\addplot [ultra thick, dodgerblue0143213, mark=*]
table {%
2 12.8
3 8.61
4 7.1
5 6.38
6 5.3
7 5.01
8 4.75
9 4.56
10 4.36
11 4.14
12 4.03
13 3.91
14 3.89
15 3.81
16 3.81
17 3.67
18 3.59
19 3.53
20 3.51
};
\addlegendentry{Ours (unbiased)}
\end{axis}

\end{tikzpicture}}}}\hfil   
\subfloat[CUB-200-2011]{\scalebox{0.61}{{
\begin{tikzpicture}

\definecolor{dodgerblue0143213}{RGB}{0,143,213}
\definecolor{goldenrod22917456}{RGB}{229,174,56}
\definecolor{lightgray203}{RGB}{203,203,203}
\definecolor{lightgray204}{RGB}{204,204,204}
\definecolor{tomato2527948}{RGB}{252,79,48}
\definecolor{whitesmoke240}{RGB}{240,240,240}
\definecolor{olivedrab10914479}{RGB}{109,144,79}

\begin{axis}[
axis background/.style={fill=white},
axis line style={whitesmoke240},
legend cell align={left},
legend style={
  fill opacity=0.8,
  draw opacity=1,
  text opacity=1,
  draw=lightgray204,
  fill=white, at={(0.97,1.2)}
},
tick align=outside,
tick pos=left,
x grid style={lightgray203},
xlabel={Number of samples},
xmajorgrids,
xmin=-0.15, xmax=25.15,
xtick style={color=black},
y grid style={lightgray203},
ylabel={MAPE (\%)},
ymajorgrids,
ymin=4.289, ymax=32.691,
ytick style={color=black},
yscale=0.8
]

\addplot [ultra thick, goldenrod22917456, mark size=1.5pt, mark=square*]
table {%
2 17.3
3 15.3
4 12.4
5 10.9
6 9.67
7 9.31
8 8.94
9 8.75
10 8.79
11 9.38
12 10
13 11.1
14 11.8
15 12.5
16 13.9
17 15.1
18 16
19 16.9
20 17.6
21 18.3
22 18.8
23 19.8
24 20.2
};
\addlegendentry{DB Index}
\addplot [ultra thick, tomato2527948, mark size=1.5pt, mark=diamond*]
table {%
2 30.6
3 22.2
4 16
5 14.2
6 12.7
7 11.7
8 10.3
9 9.41
10 8.7
11 8.31
12 8.04
13 7.5
14 7.25
15 6.95
16 6.6
17 6.49
18 6.45
19 6.36
20 6.31
21 6.1
22 6.03
23 5.93
24 5.88
};
\addlegendentry{Cross-validation}

\addplot [ultra thick, olivedrab10914479, mark=triangle*, mark size=1.5pt]
table {%
2 26
3 25.6
4 25
5 24.2
6 23.1
7 21.9
8 20.6
9 19.6
10 18.4
11 17.3
12 16.3
13 15.4
14 14.9
15 14.3
16 13.7
17 13.2
18 12.6
19 12.2
20 11.6
21 11.2
22 10.8
23 10.5
24 10.1
};
\addlegendentry{Ours (biased)}
\addplot [ultra thick, dodgerblue0143213, mark size=1.5pt, mark=*]
table {%
2 25.5
3 18.4
4 15
5 12.8
6 11.9
7 10.7
8 9.82
9 9.2
10 8.45
11 8.26
12 8.03
13 7.58
14 7.44
15 7.12
16 6.95
17 6.74
18 6.65
19 6.55
20 6.57
21 6.27
22 6.14
23 6.11
24 6.02
};
\addlegendentry{Ours (unbiased)}
\end{axis}

\end{tikzpicture}}}}\hfil
\subfloat[Describable Textures (DTD)]{\scalebox{0.61}{{
\begin{tikzpicture}

\definecolor{dodgerblue0143213}{RGB}{0,143,213}
\definecolor{goldenrod22917456}{RGB}{229,174,56}
\definecolor{lightgray203}{RGB}{203,203,203}
\definecolor{lightgray204}{RGB}{204,204,204}
\definecolor{tomato2527948}{RGB}{252,79,48}
\definecolor{whitesmoke240}{RGB}{240,240,240}
\definecolor{olivedrab10914479}{RGB}{109,144,79}

\begin{axis}[
axis background/.style={fill=white},
axis line style={whitesmoke240},
legend cell align={left},
legend style={
  fill opacity=0.8,
  draw opacity=1,
  text opacity=1,
  draw=lightgray204,
  fill=white, at={(0.97,1.2)}
},
tick align=outside,
tick pos=left,
x grid style={lightgray203},
xlabel={Number of samples},
xmajorgrids,
xmin=-1.45, xmax=52.45,
xtick style={color=black},
y grid style={lightgray203},
ylabel={MAPE (\%)},
ymajorgrids,
ytick style={color=black},
yscale=0.8
]
\addplot [ultra thick, goldenrod22917456, mark=square*, mark size=1.5pt]
table {%
2 14.5
3 11.6
4 10.5
5 9.56
6 8.21
7 7.55
8 7.17
9 6.86
10 6.62
11 6.37
12 6.25
13 6.57
14 6.67
15 6.63
16 6.83
17 7.27
18 7.68
19 8.11
20 8.54
21 8.53
22 8.91
23 9.23
24 9.52
25 9.63
26 9.89
27 10
28 10
29 10.3
30 10.9
31 11.1
32 11.3
33 11.6
34 12
35 12.1
36 12.2
37 12.3
38 12.5
39 13.1
40 13.3
41 13.6
42 13.9
43 14
44 14.2
45 14.4
46 14.6
47 15
48 15.1
49 15.3
50 15.6
};
\addlegendentry{DB Index}

\addplot [ultra thick, tomato2527948, mark=diamond*, mark size=1.5pt]
table {%
2 30.8
3 22.8
4 17.4
5 14.7
6 13.1
7 11.9
8 11.2
9 10.1
10 9.14
11 8.63
12 8.35
13 7.77
14 7.54
15 6.95
16 6.8
17 6.69
18 6.47
19 6.27
20 6.08
21 5.69
22 5.61
23 5.5
24 5.45
25 5.31
26 5.33
27 5.2
28 5.14
29 5.14
30 4.99
31 4.89
32 4.88
33 4.84
34 4.79
35 4.7
36 4.62
37 4.54
38 4.55
39 4.41
40 4.32
41 4.26
42 4.23
43 4.22
44 4.19
45 4.16
46 4.15
47 4.18
48 4.14
49 4
50 3.95
};
\addlegendentry{Cross-validation}
\addplot [ultra thick, olivedrab10914479, mark=triangle*, mark size=1.5pt]
table {%
2 21.2
3 19.3
4 18.1
5 17.8
6 17.3
7 16.7
8 16.1
9 15.4
10 14.9
11 14.4
12 13.9
13 13.3
14 12.8
15 12.3
16 11.8
17 11.6
18 11.2
19 10.9
20 10.6
21 10.3
22 10
23 9.81
24 9.58
25 9.35
26 9.18
27 8.89
28 8.72
29 8.52
30 8.31
31 8.12
32 7.96
33 7.79
34 7.67
35 7.56
36 7.42
37 7.27
38 7.17
39 7.05
40 6.95
41 6.86
42 6.77
43 6.62
44 6.6
45 6.53
46 6.37
47 6.33
48 6.26
49 6.24
50 6.19
};
\addlegendentry{Ours (biased)}
\addplot [ultra thick, dodgerblue0143213, mark=*, mark size=1.5pt]
table {%
2 30.1
3 21
4 16.6
5 14.5
6 13.3
7 12.4
8 11.1
9 10.4
10 9.84
11 9.13
12 8.64
13 8.1
14 7.84
15 7.34
16 7.22
17 7.04
18 6.84
19 6.61
20 6.36
21 6.11
22 6.03
23 5.96
24 5.8
25 5.75
26 5.72
27 5.58
28 5.47
29 5.44
30 5.35
31 5.33
32 5.32
33 5.21
34 5.18
35 5.11
36 5
37 4.89
38 4.89
39 4.86
40 4.84
41 4.77
42 4.66
43 4.66
44 4.57
45 4.51
46 4.57
47 4.53
48 4.47
49 4.37
50 4.35
};
\addlegendentry{Ours (unbiased)}
\end{axis}

\end{tikzpicture}}}}
\vspace{-0.5mm}
\caption{Prediction error of different generalization predictors over $10^3$ few-shot classification problems with 5 classes. Figures (a,b,c) are in-domain datasets and Figures (d,e,f) are cross-domain datasets. We plot the Mean-Absolute-Percentage Error against the number of samples and compare our method (unbiased) to cross-validation or Davies-Bouldin Index.}
\label{fig:bigResults}
\end{figure*}

In this section, we report the performance of our method under different settings. For each dataset, we run $10^3$ few-shot classification problems between 5 classes which is the standard number of classes in the standardized few-shot benchmarks~\cite{simpleshot}. For each run, we predict the accuracy ($1-\hat{P}_e$) and compare it to the real accuracy obtained using the large number of samples available from the original datasets. We compare our method to two other approaches. The first one is the leaving-one-out cross-validation from the few available samples. The second method is the one used by~\cite{myriam} which computes the Davies-Bouldin (DB) Index, a clustering measure of inner-class and outer-class variance. For a fair comparison, we use the validation split of the original datasets to train a linear regression for the DB Index method and apply it on the few-shot tasks. For the cross-domain datasets, we use the validation split of ImageNet.

Our main metric is the Mean Absolute Percentage Error defined as: $\text{MAPE}(\hat{P}_e, P_e)~=~\frac{\left|P_{e}-\hat{P}_{e}\right|}{1-P_{e}}$ averaged over $10^3$ problems. The results of our experiments are reported in Figure~\ref{fig:bigResults}. For each dataset\footnote{Some datasets have a limited number of samples per class (VGG-Flower and CUB-200-2011). In order to ensure that we have enough samples to measure the ground-truth accuracy, we plot less than 50 samples for these datasets.}, we plot the MAPE of the different approaches against the number of samples per class. We observe that our proposed method (Ours unbiased) outperforms cross-validation when very few labeled samples are available ($k\leq10$). The performance of cross-validation becomes more competitive when more samples are available, which is expected given the robustness of cross-validation with large data. Furthermore, the DB Index method is only competitive on Mini-ImageNet and tends to collapse in cross-domain or harder datasets. Moreover, our method without the bias correction (Ours biased) does not yield good results, showing the importance of the unbiased estimator for the distances. 

Furthermore, our method is more efficient at predicting generalization. For example on Tiered-ImageNet, when the cross-validation method would require 17 samples to estimate generalization at a certain performance, our method would only require 10 samples to match it. On other datasets such as ImageNet and DTD, the gap is smaller, however on average our method does not perform worse than the cross-validation on the tested datasets. 

\begin{figure}[!t]
    \centering
    \scalebox{1}{\input{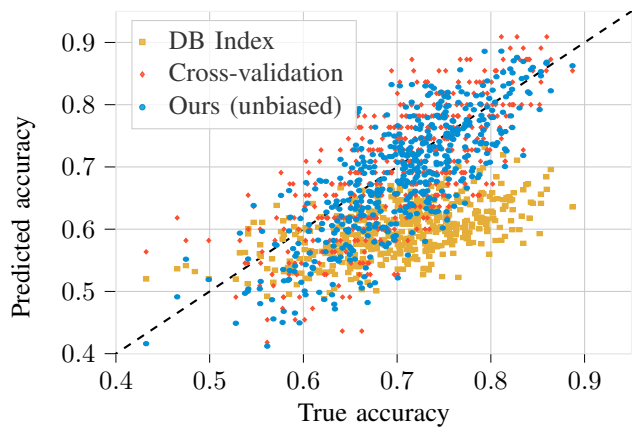}}
    \caption{Scatter plot of 5-class few-shot problems with 10 samples per class from ImageNet. Each point represents a different problem with a true ground-truth accuracy plotted against the predicted accuracy from the different methods.}
    \label{fig:resultsScatterImageNet}
\end{figure}

Figures~\ref{fig:resultsScatterImageNet} shows a scatter plot of the true accuracy and the predicted accuracy of each method for few-shot problems sampled from ImageNet. The predictions from our method are aligned with the ground truth accuracies and are less scattered than the cross-validation approach. 

\subsection{Predicting generalization as a binary classification problem} 
\begin{figure}[!htbp]
    \centering
    \scalebox{0.9}{\input{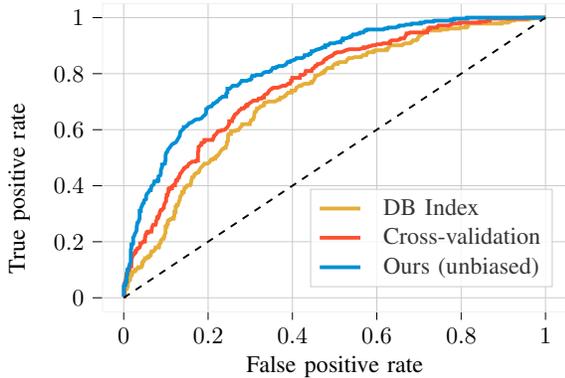}}
    \caption{Receiver operating characteristic (ROC) curve after binarizing the prediction of generalization for few-shot problems from Mini-ImageNet.}
    \label{fig:resultsROC}
\end{figure}

Let us consider now that we want to classify a problem into two classes: hard or easy. To artificially create such classes, we use a threshold of 85\% accuracy on unseen samples. We want to investigate the ability of the proposed methodology to better classify between the two classes compared to DB Index or cross-validation. We thus compute the ROC curve when generating $10^3$ few-shot problems from Mini-Imagenet. The results are depicted in Figure~\ref{fig:resultsROC}, where we see that the gap between the proposed methodology and available alternatives is even higher than in Figure~\ref{fig:bigResults}. This is because we focus here on high accuracy, where our model typically reaches the best estimation of error.

\section{Discussion and limitations}

A first observation is that the DB Index performs poorly in our experiments. We believe that this is due to the large domain gap, causing a mismatch in the learned parameters of the linear regression. We can observe this behaviour in Figure~\ref{fig:resultsScatterImageNet} where the DB Index predictions are misaligned with the ground-truth accuracies. We have experimented with different learned functions using polynomial functions with different degrees, the linear regression yielded the best results. On the other hand, our method and the cross-validation do not suffer from this behaviour. 

Furthermore, when predicting generalization from few annotated samples, all the existing techniques, except for cross-validation, are relying on a measure of clustering which depends on computing the distances and the variances of the classes, these methods could benefit from the bias correction step especially when working with few samples. The unbiased estimator is not restricted to our method. Other methods relying on the neural network gradients during training~\cite{sharpness} or the analysis of the function space defined by the network~\cite{ntk} require training the feature extractor on the few-shot task which needs a large number of samples. In case of binary classification, our method can be seen as a clustering score similar to the DB Index on which we apply a Gaussian kernel. However, the analogy only holds for binary classification. While clustering scores are either a global measure for the data distribution (DB Index) or could be used to compute pairwise probability errors between classes, our method has the advantage of estimating the class-conditional densities to compute the overlap between all the classes.
\begin{figure}[!htbp]
    \centering
\begin{tikzpicture}

\definecolor{green}{RGB}{0,128,0}
\definecolor{lightgray203}{RGB}{203,203,203}
\definecolor{lightgray204}{RGB}{204,204,204}
\definecolor{orange}{RGB}{255,165,0}
\definecolor{whitesmoke240}{RGB}{240,240,240}
\definecolor{dodgerblue0143213}{RGB}{0,143,213}
\definecolor{goldenrod22917456}{RGB}{229,174,56}
\definecolor{tomato2527948}{RGB}{252,79,48}
\definecolor{olivedrab10914479}{RGB}{109,144,79}

\begin{axis}[
axis background/.style={fill=white},
axis line style={whitesmoke240},
legend cell align={left},
legend style={fill opacity=0.8, draw opacity=1, text opacity=1, draw=lightgray204, fill=white, at={(0.97,1.2)}},
tick align=outside,
tick pos=left,
title={},
x grid style={lightgray203},
xlabel={\(\displaystyle SNR_{DB}\)},
xmajorgrids,
xmin=-0.389075636863708, xmax=8.17058837413788,
xtick style={color=black},
y grid style={lightgray203},
ylabel={MAPE (\%)},
ymajorgrids,
ymin=-3.70732051699015, ymax=77.8647584151506,
ytick style={color=black},
yscale=0.8
]
\addplot [ultra thick, tomato2527948, mark=diamond*, mark size=1.5pt]
table {%
0 19.8549633026123
0.690809369087219 17.9554080963135
1.28666627407074 17.5923385620117
1.81054663658142 16.1146259307861
2.27798080444336 14.7169370651245
2.69995784759521 13.7356338500977
3.08454012870789 11.8718166351318
3.43781971931458 10.9973926544189
3.76451110839844 10.0930004119873
4.06833744049072 8.75243282318115
4.35229110717773 7.41024017333984
4.61881256103516 6.26507186889648
4.8699197769165 5.22975635528564
5.10729837417603 4.35409545898438
5.33237218856812 3.88631224632263
5.54635334014893 3.11688995361328
5.75028467178345 2.3518123626709
5.945068359375 1.80735123157501
6.13148975372314 1.20720946788788
6.31023693084717 0.756788015365601
6.48191738128662 0.462376922369003
6.64706754684448 0.316720008850098
6.80616807937622 0.169514030218124
6.95964431762695 0.0901266038417816
7.10788249969482 0.0438233576714993
7.25122690200806 0.0235212948173285
7.38999128341675 0.0197619013488293
7.52445793151855 0.0025075227022171
7.6548867225647 0.000751878949813545
7.78151273727417 0.000501252652611583
};
\addlegendentry{Cross-validation}
\addplot [ultra thick, olivedrab10914479, mark=triangle*, mark size=1.5pt]
table {%
0 74.1569366455078
0.690809369087219 66.4339599609375
1.28666627407074 58.5050506591797
1.81054663658142 50.917839050293
2.27798080444336 43.4948043823242
2.69995784759521 36.3544464111328
3.08454012870789 29.7746334075928
3.43781971931458 24.4911823272705
3.76451110839844 19.6709461212158
4.06833744049072 15.1833639144897
4.35229110717773 11.7595453262329
4.61881256103516 8.66726398468018
4.8699197769165 6.55400228500366
5.10729837417603 4.6643180847168
5.33237218856812 3.2817964553833
5.54635334014893 2.20067453384399
5.75028467178345 1.50390028953552
5.945068359375 0.962960720062256
6.13148975372314 0.615659177303314
6.31023693084717 0.369266480207443
6.48191738128662 0.194363281130791
6.64706754684448 0.130057901144028
6.80616807937622 0.0624638125300407
6.95964431762695 0.0406453236937523
7.10788249969482 0.0198282636702061
7.25122690200806 0.00854630023241043
7.38999128341675 0.00476690102368593
7.52445793151855 0.00252252514474094
7.6548867225647 0.000751878949813545
7.78151273727417 0.000506253505591303
};
\addlegendentry{Ours (biased)}
\addplot [ultra thick, dodgerblue0143213, mark=*, mark size=1.5pt]
table {%
0 14.2276554107666
0.690809369087219 12.5535202026367
1.28666627407074 11.9511404037476
1.81054663658142 11.9381843185425
2.27798080444336 11.9226341247559
2.69995784759521 11.098783493042
3.08454012870789 10.2041015625
3.43781971931458 9.7119083404541
3.76451110839844 8.6392765045166
4.06833744049072 7.54780006408691
4.35229110717773 6.18725252151489
4.61881256103516 4.68100357055664
4.8699197769165 3.86787509918213
5.10729837417603 2.88758778572083
5.33237218856812 2.49332761764526
5.54635334014893 1.77929127216339
5.75028467178345 1.37087786197662
5.945068359375 1.09383511543274
6.13148975372314 0.736416578292847
6.31023693084717 0.5665642619133
6.48191738128662 0.388852030038834
6.64706754684448 0.303916215896606
6.80616807937622 0.219856023788452
6.95964431762695 0.144908636808395
7.10788249969482 0.0947575271129608
7.25122690200806 0.063512809574604
7.38999128341675 0.0410765111446381
7.52445793151855 0.0300625171512365
7.6548867225647 0.0194020345807076
7.78151273727417 0.0133264428004622
};
\addlegendentry{Ours (unbiased)}

\addplot [ultra thick, brown]
table {%
0 3.32558298110962
0.690809369087219 3.41114735603333
1.28666627407074 3.10629892349243
1.81054663658142 2.94476366043091
2.27798080444336 2.78219604492188
2.69995784759521 2.5068199634552
3.08454012870789 2.26372909545898
3.43781971931458 2.16438484191895
3.76451110839844 1.98385679721832
4.06833744049072 1.77506601810455
4.35229110717773 1.52775514125824
4.61881256103516 1.32947397232056
4.8699197769165 1.19922196865082
5.10729837417603 0.959894180297852
5.33237218856812 0.803239822387695
5.54635334014893 0.671056687831879
5.75028467178345 0.52955150604248
5.945068359375 0.424422800540924
6.13148975372314 0.323393017053604
6.31023693084717 0.254369676113129
6.48191738128662 0.17505943775177
6.64706754684448 0.143244475126266
6.80616807937622 0.0885513573884964
6.95964431762695 0.0586590804159641
7.10788249969482 0.0309360213577747
7.25122690200806 0.0143413664773107
7.38999128341675 0.00768721895292401
7.52445793151855 0.00386268552392721
7.6548867225647 0.00128196692094207
7.78151273727417 0.000771284976508468
};
\addlegendentry{Oracle}

\end{axis}

\end{tikzpicture}
    \caption{Prediction error on $10^4$ binary 10-shot classification problems generated from isotropic Gaussian distributions with different SNR values. The oracle model is using our approach with the true parameter distributions.}
    \label{fig:resultsSynthetic}
\end{figure}
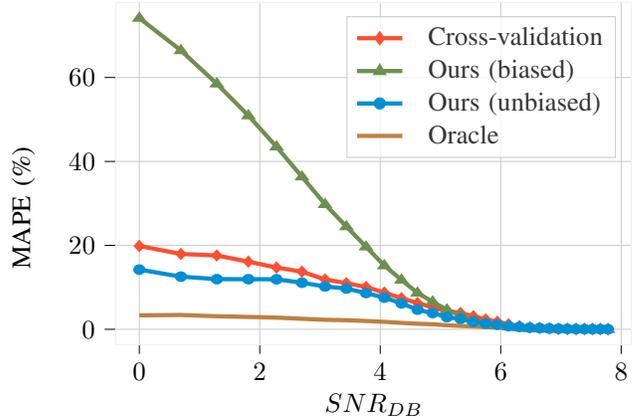

Predicting error depends on the accuracy of the few-shot problems. Hard few-shot problems have a low SNR which makes the estimation of accuracy more difficult. On the other hand, good separation between the classes in the latent space leads to a better estimation of generalization. This explains the better performance on in-domain datasets compared to cross-domain, as well as the results of Figure~\ref{fig:resultsROC}. We run an experiment in Figure~\ref{fig:resultsSynthetic} to show the effect of SNR on the performance of both our method and cross-validation with 10 samples per class. The performance drastically drops in low SNR regimes especially for our method without correcting for the bias. This is expected since the bias is higher when the noise is high. We also include the performance of an oracle model, which is using our method with the true parameters of underlying distribution of the data. As the SNR value gets lower, the separation between the Gaussian distributions becomes more difficult which also impacts the estimation of the accuracy. 

\begin{figure}[!htbp]
    \centering
    \scalebox{0.9}{\input{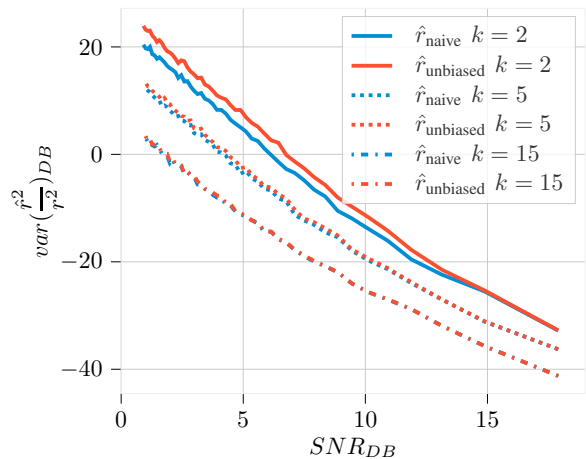}}
    \caption{Variance of both estimators against different SNR values in log-log scale. We normalize the estimates by the real distances. The unbiased estimator has a slightly higher variance than the naive estimator. The variance of the two estimators are asymptotically of $\mathcal{O}(\frac{1}{\text{SNR}})$.}
    \label{fig:resultsVarianceEstimator}
\end{figure}

Finally, although the proposed estimator is unbiased, it still suffers from a large variance. We show this in Figure~\ref{fig:resultsVarianceEstimator} where the variance of the estimator is slightly higher than the naive estimator. The variance of both estimators are particularly high in low SNR regimes. 

\section{Conclusion}
We propose a model based approach to estimate the generalization capability of few-shot classifiers and provide statistical performance bounds. Our method outperforms the leave-one-out cross-validation and the Davis-Bouldin score-based estimator for different Signal-to-Noise regimes and a small number of labeled samples. This is especially important in the few-shot context. Our method strongly relies on unbiased estimates of the inter-class distances, which is a key contribution of this paper. Note that our method can be generalized to transfer-based few-shot learners with any distance-based classifier. Although we improve upon existing methods, we think that it opens up interesting new directions for further research.

\bibliographystyle{IEEEtran}
\bibliography{bib}
\end{document}


\onecolumn
\title{A Statistical Model for Predicting Generalization in Few-Shot Classification\\\scalebox{0.65}{Supplementary Materials}}

\author{\IEEEauthorblockN{Yassir Bendou\Mark{1}, Vincent Gripon\Mark{1}, Bastien Pasdeloup\Mark{1}, Giulia Lioi\Mark{1}, Lukas Mauch\Mark{2}, \\Stefan Uhlich\Mark{2}, Fabien Cardinaux\Mark{2}, Ghouthi Boukli Hacene\Mark{2}\Mark{3} and Javier Alonso Garcia\Mark{2}} \\
\IEEEauthorblockA{\Mark{1}IMT Atlantique, Lab-STICC, UMR CNRS 6285, F-29238 Brest, France} 
\IEEEauthorblockA{\Mark{2}Sony Europe, R\&D Center, Stuttgart Laboratory 1, Germany}
\IEEEauthorblockA{\Mark{3}Mila, Montréal, Canada}
\IEEEauthorblockA{\Mark{1}name.surname@imt-atlantique.fr, \Mark{2}name.surname@sony.com}}
\maketitle
\section{Statistical bound for the probability estimate}
\begin{lemma}
\label{lemma:lemma1}
Given two classes represented by two univariate isotropic distributions $\mathcal{N}_a(\mu_a, \sigma^2)$ and $\mathcal{N}_b(\mu_b, \sigma^2)$ with shared and known standard deviations, the true probability of error $P_e$ is bounded by: 
\begin{equation}
  P\left(\left|P_e-\hat{P_e}\right| \leq \frac{\phi^{'}(0)}{\sqrt{2k}}\left|\phi^{-1}\left(1-\frac{\alpha}{2}\right)\right|\right)\geq1-\alpha\;,  
\end{equation}

with a probability $1-\alpha$, where $\hat{P}_e$ is the probability error estimate from a sequence of $k$ i.i.d random variables from the two distributions with empirical mean estimates $\hat{\mu}_a$ and $\hat{\mu}_b$ , $\phi$ is the cumulative distribution function of $\mathcal{N}(0,1)$, $\phi^{-1}$ its inverse and $\phi^{'}$ its derivative, with $\phi^{'}(0)=\frac{1}{\sqrt{2\pi}}$.
\end{lemma}
\begin{myproof}{Lemma}{\ref{lemma:lemma1}}
Let $(a_1, a_2, \cdots, a_{k})$ and $(b_1, b_2, \cdots, b_{k})$ be two sequences of i.i.d random variables drawn from two respective univariate normal distributions $\mathcal{N}\left(\mu_a, \sigma^2\right)$ and $\mathcal{N}\left(\mu_b, \sigma^2\right)$. Let $\hat{\mu}_a=\frac{1}{k} \sum_{i=1}^{k} a_m$ and $\hat{\mu}_b=\frac{1}{k} \sum_{i=1}^{k} b_m$ be the empirical mean estimates of each of the two sequences. Then $\hat{\mu}_a-\hat{\mu}_b\sim\mathcal{N}(\mu_a-\mu_b, \frac{2\sigma^2}{k})$. 
Therefore, $\hat{t}~=~\frac{\left(\hat{\mu}_a-\hat{\mu}_b\right)-\left(\mu_a-\mu_b\right)}{\sigma\sqrt{2/k}} \sim \mathcal{N}\left(0, 1\right)$. With a probability $1-\alpha$, the confidence bound for the standard normal is: 

$$P\left(\left|\frac{(\hat{\mu}_a-\hat{\mu}_b)-(\mu_a-\mu_b)}{\sigma\sqrt{2/k}}\right|\leq|\phi^{-1}\left(1-\alpha/2\right)|\right)=1-\alpha\;.$$
We also know that :
$$\left|P_e-\hat{P_e}\right| = \left|\phi\left(\frac{\abs{\mu_a-\mu_b}}{2\sigma}\right)-\phi\left(\frac{\abs{\hat{\mu}_a-\hat{\mu}_b}}{2\sigma}\right)\right|\;.$$

Given that $\phi$ is a differentiable function then it is Lipschitz continuous and since $\forall x\in \mathbb{R}: \phi^{'}(x)\leq \phi^{'}(0)$, then: 
$$\forall (x,y)\in\mathbb{R}^2: \left|\phi(x)-\phi(y)\right|\leq\phi^{'}(0)\left|x-y\right|\;.$$
Using this property and the triangular inequality, we get: 
$$\left|P_e-\hat{P_e}\right| \leq\phi^{'}(0) \left|\frac{\left|\mu_a-\mu_b\right|}{2\sigma}-\frac{\left|\hat{\mu}_a-\hat{\mu}_b\right|}{2\sigma}\right|\;.$$
$$\frac{\sqrt{2k}\left|P_e-\hat{P_e}\right|}{\phi^{'}(0)} \leq \left|\frac{\left(\mu_a-\mu_b\right)-\left(\hat{\mu}_a-\hat{\mu}_b\right)}{\sigma\sqrt{2/k}}\right|\;.$$ 
Hence with a probability $1-\alpha$: 

$$\frac{\sqrt{2k}\left|P_e-\hat{P_e}\right|}{\phi^{'}(0)} \leq \left|\frac{\left(\mu_a-\mu_b\right)-\left(\hat{\mu}_a-\hat{\mu}_b\right)}{\sigma\sqrt{2/k}}\right|\leq|\phi^{-1}\left(1-\alpha/2\right)|\;.$$ 
Finally, since density functions are positive, we can rewrite with a probability $1-\alpha$:
$$P\left(\left|P_e-\hat{P_e}\right| \leq \frac{\phi^{'}\left(0\right)}{\sqrt{2k}}\left|\phi^{-1}\left(1-\alpha/2\right)\right|\right)\geq1-\alpha\;.$$
\end{myproof}

\begin{figure}[htbp]
\centering
\subfloat[2-class, $R^2=0.9979$]{\scalebox{0.8}{{
\begin{tikzpicture}

\definecolor{dodgerblue0143213}{RGB}{0,143,213}
\definecolor{lightgray203}{RGB}{203,203,203}
\definecolor{lightgray204}{RGB}{204,204,204}
\definecolor{whitesmoke240}{RGB}{240,240,240}

\begin{axis}[
axis background/.style={fill=white},
axis line style={whitesmoke240},
legend cell align={left},
legend style={
  fill opacity=0.8,
  draw opacity=1,
  text opacity=1,
  at={(0.03,0.97)},
  anchor=north west,
  draw=lightgray204,
  fill=white
},
tick align=outside,
tick pos=left,
title={},
x grid style={lightgray203},
xlabel={number of samples per class},
xmajorgrids,
xmin=-1.45, xmax=52.45,
xtick style={color=black},
y grid style={lightgray203},
ymajorgrids,
ymin=-330.127980452481, ymax=4388.97131008107,
ytick style={color=black}
]
\addplot [draw=dodgerblue0143213, fill=dodgerblue0143213, mark=*, only marks]
table{%
x  y
1 38.2048873901367
2 88.5295562744141
3 141.928634643555
4 201.904525756836
5 278.678985595703
6 351.559112548828
7 414.934967041016
8 484.972778320312
9 559.335571289062
10 633.683654785156
11 705.676330566406
12 792.9677734375
13 874.937194824219
14 944.045837402344
15 1032.70178222656
16 1142.24255371094
17 1241.72265625
18 1316.5810546875
19 1399.20727539062
20 1468.65698242188
21 1560.83959960938
22 1666.53784179688
23 1757.98999023438
24 1866.22106933594
25 1944.83215332031
26 1970.57861328125
27 2025.49133300781
28 2138.89672851562
29 2231.4990234375
30 2371.50170898438
31 2434.36596679688
32 2534.07690429688
33 2593.15502929688
34 2665.14086914062
35 2768.64501953125
36 2849.84692382812
37 3000.0859375
38 2993.666015625
39 3086.755859375
40 3253.287109375
41 3318.21606445312
42 3416.4560546875
43 3538.5517578125
44 3567.04418945312
45 3692.81689453125
46 3812.61767578125
47 3901.2333984375
48 3976.5380859375
49 4107.2578125
50 4174.466796875
};
\addlegendentry{$\frac{1}{\left|P_e-\hat{P}_e\right|^2}$}
\addplot [ultra thick, red]
table {%
1 -115.62346724641
2 -29.8175424271822
3 55.9883823920458
4 141.794307211274
5 227.600232030502
6 313.40615684973
7 399.212081668958
8 485.018006488186
9 570.823931307414
10 656.629856126642
11 742.43578094587
12 828.241705765098
13 914.047630584326
14 999.853555403554
15 1085.65948022278
16 1171.46540504201
17 1257.27132986124
18 1343.07725468047
19 1428.88317949969
20 1514.68910431892
21 1600.49502913815
22 1686.30095395738
23 1772.10687877661
24 1857.91280359583
25 1943.71872841506
26 2029.52465323429
27 2115.33057805352
28 2201.13650287275
29 2286.94242769197
30 2372.7483525112
31 2458.55427733043
32 2544.36020214966
33 2630.16612696889
34 2715.97205178811
35 2801.77797660734
36 2887.58390142657
37 2973.3898262458
38 3059.19575106503
39 3145.00167588425
40 3230.80760070348
41 3316.61352552271
42 3402.41945034194
43 3488.22537516117
44 3574.0312999804
45 3659.83722479962
46 3745.64314961885
47 3831.44907443808
48 3917.25499925731
49 4003.06092407654
50 4088.86684889576
};
\end{axis}

\end{tikzpicture}}}}\hfil 
\subfloat[5-class, $R^2=0.9981$]{\scalebox{0.8}{{
\begin{tikzpicture}

\definecolor{dodgerblue0143213}{RGB}{0,143,213}
\definecolor{lightgray203}{RGB}{203,203,203}
\definecolor{lightgray204}{RGB}{204,204,204}
\definecolor{whitesmoke240}{RGB}{240,240,240}

\begin{axis}[
axis background/.style={fill=white},
axis line style={whitesmoke240},
legend cell align={left},
legend style={
  fill opacity=0.8,
  draw opacity=1,
  text opacity=1,
  at={(0.03,0.97)},
  anchor=north west,
  draw=lightgray204,
  fill=white
},
tick align=outside,
tick pos=left,
title={},
x grid style={lightgray203},
xlabel={number of samples per class},
xmajorgrids,
xmin=-1.45, xmax=52.45,
xtick style={color=black},
y grid style={lightgray203},
ymajorgrids,
ymin=-152.265233156868, ymax=6458.19541509307,
ytick style={color=black}
]
\addplot [draw=dodgerblue0143213, fill=dodgerblue0143213, mark=*, only marks]
table{%
x  y
1 148.210250854492
2 245.667602539062
3 363.009338378906
4 468.463073730469
5 604.975769042969
6 715.122680664062
7 820.6953125
8 930.604675292969
9 1042.44909667969
10 1164.90466308594
11 1284.3896484375
12 1428.98449707031
13 1572.91979980469
14 1688.51110839844
15 1839.705078125
16 2002.13146972656
17 2161.39965820312
18 2278.2294921875
19 2381.51098632812
20 2538.63598632812
21 2696.12841796875
22 2816.71166992188
23 2931.47778320312
24 3102.71850585938
25 3192.09130859375
26 3321.86840820312
27 3426.45092773438
28 3577.93286132812
29 3621.33618164062
30 3729.17456054688
31 3869.93383789062
32 3991.37426757812
33 4175.576171875
34 4296.0068359375
35 4385.61328125
36 4555.80322265625
37 4591.1572265625
38 4727.36865234375
39 4813.341796875
40 4949.76171875
41 5060.3173828125
42 5166.84326171875
43 5262.619140625
44 5341.39306640625
45 5599.59521484375
46 5581.94189453125
47 5708.3408203125
48 5790.7626953125
49 5805.2939453125
50 5982.9462890625
};
\addlegendentry{$\frac{1}{\left|P_e-\hat{P}_e\right|^2}$}
\addplot [ultra thick, red]
table {%
1 152.376138284159
2 274.934174871864
3 397.492211459569
4 520.050248047274
5 642.608284634979
6 765.166321222684
7 887.724357810389
8 1010.28239439809
9 1132.8404309858
10 1255.3984675735
11 1377.95650416121
12 1500.51454074891
13 1623.07257733662
14 1745.63061392432
15 1868.18865051203
16 1990.74668709973
17 2113.30472368744
18 2235.86276027514
19 2358.42079686285
20 2480.97883345055
21 2603.53687003826
22 2726.09490662596
23 2848.65294321367
24 2971.21097980137
25 3093.76901638908
26 3216.32705297678
27 3338.88508956449
28 3461.44312615219
29 3584.0011627399
30 3706.5591993276
31 3829.11723591531
32 3951.67527250301
33 4074.23330909072
34 4196.79134567842
35 4319.34938226613
36 4441.90741885383
37 4564.46545544154
38 4687.02349202924
39 4809.58152861695
40 4932.13956520465
41 5054.69760179236
42 5177.25563838006
43 5299.81367496777
44 5422.37171155547
45 5544.92974814318
46 5667.48778473088
47 5790.04582131859
48 5912.60385790629
49 6035.161894494
50 6157.7199310817
};
\end{axis}

\end{tikzpicture}}}}\hfil 
\caption{Inverse quadratic difference between the estimate of the probability of error and the true probability of error. For each point, we average $10^3$ few-shot problems from Mini-ImageNet for both binary and 5-class problems.}
\label{fig:resultslemma1}
\end{figure}
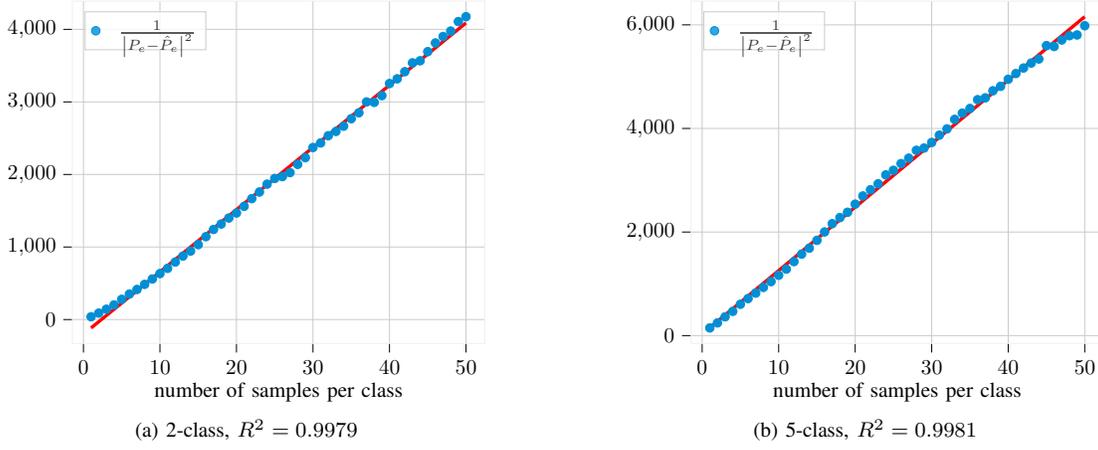

Figure~\ref{fig:resultslemma1} shows the $\frac{1}{\sqrt{k}}$ behaviour of the confidence bound on real data. We generate $10^3$ few-shot problems and plot the average difference between the true probability error and the estimated probability error using $\frac{1}{\left|P_e-\hat{P}_e\right|^2}$ against the number of samples $k$. We run our experiment with both binary classification and 5-class problems and we fit a linear regression for each of the configurations with respective coefficients of determination of $R^2=0.9979$ and $R^2=0.9981$.

\section{Bias of the naive distance estimate}
\begin{lemma}
\label{lemma:lemma2}
Let $(\bm{a}_1, \bm{a}_2, \cdots, \bm{a}_{k})$ and $(\bm{b}_1, \bm{b}_2, \cdots, \bm{b}_{k})$ be two sequences of i.i.d random variables drawn from their respective multivariate probability distributions $p_{a}$ and $p_{b}$ assumed independent with finite expected values $\bm{\mu}_a$ and $\bm{\mu}_b$ and finite second order moment with covariance matrices $\mathbf{\Sigma}_{a}$ and $\bm{\Sigma}_{b}$. Let $\hat{r}$ be the naive estimator for the distances using $\hat{\bm{\mu}}_{a}~=~\frac{1}{k} \sum_{i=1}^{k} \bm{a}_i$ and $\bm{\hat{\mu}}_{b}~=~\frac{1}{k} \sum_{i=1}^{k} \bm{b}_i$ the mean estimator of each of the two sequences, then:

\begin{equation}
    \mathbb{E}_{\substack{\bm{a}\sim p_a \\ \bm{b}\sim p_b}}\left(\hat{r}^2\right) - r^2 =  \frac{\Tr\left(\mathbf{\Sigma}_{a}+\mathbf{\Sigma}_{b}\right)}{k}\;.
\end{equation}
\end{lemma}

\begin{myproof}{Lemma}{\ref{lemma:lemma2}}
For this proof, we exploit the independence of the variables and the second order definition: $\mathbb{E}_{\bm{a}\sim p_a}\left[\bm{a}\bm{a}^T\right]=\mathbf{\Sigma}_{a}+\mathbb{E}_{\bm{a}\sim p_a}\left[\bm{a}\right]\mathbb{E}_{\bm{a}\sim p_a}\left[\bm{a}\right]^T$.
\begin{eqnarray*}
\lefteqn{\mathbb{E}_{\substack{\bm{a}\sim p_{a} \\ \bm{b}\sim p_{b}}}\left[\norm{\hat{\bm{\mu}}_{\bm{a}}-\hat{\bm{\mu}}_{\bm{b}}}^2_2\right] = \frac{1}{k^2}\mathbb{E}_{\substack{\bm{a}\sim p_{a} \\ \bm{b}\sim p_{b}}}\left[\norm{\sum_{i=1}^{k}\left(\bm{a}_i-\bm{b}_i\right)}^2_2\right]} \\
&=& \frac{1}{k^2}\mathbb{E}_{\substack{\bm{a}\sim p_{a} \\ \bm{b}\sim p_{b}}}\left[\Tr\left(\sum_{i=1}^{k} \sum_{j=1}^{k}\left(\bm{a}_i-\bm{b}_i\right)\left(\bm{a}_j-\bm{b}_j\right)^T\right)\right] \\
&=& \frac{1}{k^2}\sum_{\substack{i, j\\
   i\neq j}}^{k}\Tr\left(\mathbb{E}_{\substack{\bm{a}_i\sim p_{a} \\ \bm{b}_i\sim p_{b}}}\left(\bm{a}_i-\bm{b}_i\right)\mathbb{E}_{\substack{\bm{a}_j\sim p_{a} \\ \bm{b}_j\sim p_{b}}}\left(\bm{a}_j-\bm{b}_j\right)^T\right) +  \frac{1}{k^2}\sum_{i=1}^{k}\Tr\left(\mathbb{E}_{\substack{\bm{a}_i\sim p_{a} \\ \bm{b}_i\sim p_{b}}}\left[\left(\bm{a}_i-\bm{b}_i\right)\left(\bm{a}_i-\bm{b}_i\right)^T\right]\right) \\
&=& \frac{k^2-k}{k^2}\norm{\bm{\mu}_{\bm{a}}-\bm{\mu}_{\bm{b}}}^2_2 + \frac{1}{k}\Tr\left(\mathbb{E}_{\bm{a}\sim p_{a}}\left[\bm{a}\bm{a}^T\right]\right)+  \frac{1}{k}\Tr\left(\mathbb{E}_{\bm{b}\sim p_{b}}\left[\bm{b}\bm{b}^T\right]\right)- \frac{2}{k}\Tr\left(\mathbb{E}_{\substack{\bm{a}\sim p_{a}\\\bm{b}\sim p_{b}}}\left[\bm{a}\bm{b}^T\right]\right) \\
&=& \left(1-\frac{1}{k}\right)\norm{\bm{\mu}_{\bm{a}}-\bm{\mu}_{\bm{b}}}^2_2 + \frac{1}{k}\Tr\left(\mathbf{\Sigma}_{a}+\mathbf{\Sigma}_{b}+\mathbb{E}_{\bm{a}\sim p_a}\left[\bm{a}\right]\mathbb{E}_{\bm{a}\sim p_a}\left[\bm{a}\right]^T + \mathbb{E}_{\bm{b}\sim p_b}\left[\bm{b}\right]\mathbb{E}_{\bm{b}\sim p_b}\left[\bm{b}\right]^T- \mathbb{E}_{\bm{a}\sim p_a}\left[\bm{a}\right]\mathbb{E}_{\bm{b}\sim p_b}\left[\bm{b}\right]^T\right) \\
&=& \norm{\bm{\mu}_{\bm{a}}-\bm{\mu}_{\bm{b}}}^2_2 + \frac{1}{k}\Tr\left(\mathbf{\Sigma}_{a}+\mathbf{\Sigma}_{b}\right)\;.
\end{eqnarray*}
\end{myproof}

\section{Additional experiments for predicting generalization}
\begin{figure}[htbp]
\centering
\subfloat[Mini-ImageNet]{\scalebox{0.65}{{
\begin{tikzpicture}

\definecolor{dodgerblue0143213}{RGB}{0,143,213}
\definecolor{goldenrod22917456}{RGB}{229,174,56}
\definecolor{lightgray203}{RGB}{203,203,203}
\definecolor{tomato2527948}{RGB}{252,79,48}
\definecolor{whitesmoke240}{RGB}{240,240,240}
\definecolor{olivedrab10914479}{RGB}{109,144,79}

\begin{axis}[
axis background/.style={fill=white},
axis line style={whitesmoke240},
tick align=outside,
tick pos=left,
title={},
legend cell align={left},
legend style={fill opacity=0.8, draw opacity=1, text opacity=1, draw=lightgray203, fill=white, at={(0.97,1.2)}},
x grid style={lightgray203},
xlabel={Number of samples},
xmajorgrids,
xmin=-1.45, xmax=52.45,
xtick style={color=black},
y grid style={lightgray203},
ylabel={MAPE (\%)},
ymajorgrids,
ymin=0, ymax=28,
ytick style={color=black},
yscale=0.8
]
\addplot [ultra thick, goldenrod22917456, mark size=1.5pt, mark=square*]
table {%
2 9.75
3 7.41
4 6.05
5 5.44
6 4.98
7 4.57
8 4.18
9 3.91
10 3.65
11 3.5
12 3.33
13 3.18
14 3.06
15 2.98
16 2.89
17 2.81
18 2.75
19 2.69
20 2.63
21 2.56
22 2.52
23 2.5
24 2.45
25 2.44
26 2.38
27 2.33
28 2.3
29 2.27
30 2.22
31 2.21
32 2.18
33 2.16
34 2.1
35 2.09
36 2.07
37 2.05
38 2.04
39 2.03
40 2.02
41 2.01
42 2
43 1.99
44 1.96
45 1.94
46 1.93
47 1.92
48 1.9
49 1.9
50 1.89
};
\addlegendentry{DB Index}
\addplot [ultra thick, tomato2527948, mark size=1.5pt, mark=diamond*]
table {%
2 24.5
3 14.8
4 10.1
5 8.13
6 6.76
7 6.11
8 5.33
9 4.81
10 4.49
11 4.19
12 3.87
13 3.7
14 3.45
15 3.37
16 3.22
17 3.08
18 3.02
19 2.91
20 2.85
21 2.68
22 2.61
23 2.57
24 2.52
25 2.44
26 2.37
27 2.3
28 2.27
29 2.21
30 2.16
31 2.16
32 2.11
33 2.1
34 2.02
35 2.02
36 1.99
37 1.95
38 1.9
39 1.86
40 1.84
41 1.81
42 1.78
43 1.75
44 1.73
45 1.72
46 1.72
47 1.73
48 1.71
49 1.7
50 1.67
};
\addlegendentry{Cross-validation}

\addplot [ultra thick, olivedrab10914479, mark size=1.5pt, mark=triangle]
table {%
2 66.9
3 50.1
4 41.6
5 36.8
6 33.4
7 31
8 29
9 27.5
10 26.1
11 25.1
12 24.1
13 23.3
14 22.6
15 22
16 21.5
17 21
18 20.5
19 20.1
20 19.7
21 19.4
22 19.1
23 18.8
24 18.5
25 18.3
26 18
27 17.8
28 17.6
29 17.4
30 17.2
31 17.1
32 16.9
33 16.8
34 16.6
35 16.5
36 16.4
37 16.2
38 16.1
39 16
40 15.9
41 15.8
42 15.7
43 15.6
44 15.5
45 15.4
46 15.3
47 15.3
48 15.2
49 15.1
50 15.1
};
\addlegendentry{Ours (biased)}

\addplot [ultra thick, dodgerblue0143213, mark size=1.5pt, mark size=1.5pt, mark=*]
table {%
2 10.4
3 7.4
4 5.29
5 4.65
6 4.03
7 3.5
8 3.07
9 2.79
10 2.52
11 2.34
12 2.19
13 2.07
14 2.01
15 1.94
16 1.88
17 1.85
18 1.79
19 1.75
20 1.72
21 1.67
22 1.63
23 1.61
24 1.59
25 1.56
26 1.53
27 1.49
28 1.46
29 1.43
30 1.4
31 1.4
32 1.39
33 1.38
34 1.36
35 1.35
36 1.34
37 1.33
38 1.31
39 1.3
40 1.29
41 1.27
42 1.27
43 1.26
44 1.25
45 1.25
46 1.25
47 1.24
48 1.22
49 1.21
50 1.21
};
\addlegendentry{Ours unbiased}
\end{axis}
\end{tikzpicture}}}}\hfil 
\subfloat[Tiered-ImageNet]{\scalebox{0.65}{{
\begin{tikzpicture}

\definecolor{dodgerblue0143213}{RGB}{0,143,213}
\definecolor{goldenrod22917456}{RGB}{229,174,56}
\definecolor{lightgray203}{RGB}{203,203,203}
\definecolor{lightgray204}{RGB}{204,204,204}
\definecolor{tomato2527948}{RGB}{252,79,48}
\definecolor{whitesmoke240}{RGB}{240,240,240}
\definecolor{olivedrab10914479}{RGB}{109,144,79}

\begin{axis}[
axis background/.style={fill=white},
axis line style={whitesmoke240},
legend cell align={left},
legend style={fill opacity=0.8, draw opacity=1, text opacity=1, draw=lightgray204, fill=white, at={(0.97,1.2)}},
tick align=outside,
tick pos=left,
x grid style={lightgray203},
xlabel={Number of samples},
xmajorgrids,
xmin=-1.45, xmax=52.45,
xtick style={color=black},
y grid style={lightgray203},
ylabel={MAPE (\%)},
ymajorgrids,
ymin=0, ymax=21.3405,
ytick style={color=black},
yscale=0.8
]
\addplot [ultra thick, goldenrod22917456, mark size=1.5pt, mark=square*]
table {%
2 9.45
3 8.05
4 7.12
5 6.53
6 6.11
7 5.64
8 5.37
9 5.09
10 4.92
11 4.71
12 4.54
13 4.44
14 4.34
15 4.22
16 4.13
17 4.04
18 3.97
19 3.89
20 3.82
21 3.77
22 3.69
23 3.64
24 3.59
25 3.56
26 3.52
27 3.48
28 3.44
29 3.39
30 3.34
31 3.3
32 3.3
33 3.28
34 3.26
35 3.25
36 3.22
37 3.2
38 3.18
39 3.15
40 3.12
41 3.09
42 3.07
43 3.05
44 3.05
45 3.04
46 3.02
47 3
48 2.98
49 2.97
50 2.95
};
\addlegendentry{DB Index}
\addplot [ultra thick, tomato2527948, mark size=1.5pt, mark=diamond*]
table {%
2 21.9
3 12
4 8.63
5 6.89
6 5.94
7 5.09
8 4.41
9 4.13
10 3.76
11 3.61
12 3.4
13 3.23
14 3.22
15 3.02
16 2.9
17 2.9
18 2.7
19 2.63
20 2.56
21 2.43
22 2.37
23 2.32
24 2.3
25 2.28
26 2.19
27 2.11
28 2.12
29 2.06
30 2.05
31 1.98
32 1.98
33 1.93
34 1.89
35 1.84
36 1.83
37 1.79
38 1.74
39 1.72
40 1.7
41 1.65
42 1.63
43 1.61
44 1.6
45 1.57
46 1.53
47 1.54
48 1.51
49 1.5
50 1.48
};
\addlegendentry{Cross-validation}

\addplot [ultra thick, olivedrab10914479, mark size=1.5pt, mark=triangle]
table {%
2 53.1
3 40.4
4 34.4
5 30.8
6 28.3
7 26.6
8 25.2
9 24.1
10 23.2
11 22.4
12 21.8
13 21.2
14 20.7
15 20.2
16 19.8
17 19.5
18 19.1
19 18.9
20 18.6
21 18.3
22 18.1
23 17.9
24 17.7
25 17.5
26 17.3
27 17.2
28 17
29 16.8
30 16.7
31 16.6
32 16.4
33 16.3
34 16.2
35 16.1
36 16
37 15.9
38 15.8
39 15.7
40 15.6
41 15.6
42 15.5
43 15.4
44 15.3
45 15.3
46 15.2
47 15.1
48 15.1
49 15
50 14.9
};
\addlegendentry{Ours (biased)}

\addplot [ultra thick, dodgerblue0143213, mark size=1.5pt, mark size=1.5pt, mark=*]
table {%
2 8.48
3 6.52
4 5.69
5 5.01
6 4.58
7 4.23
8 3.96
9 3.75
10 3.55
11 3.43
12 3.27
13 3.19
14 3.12
15 3.06
16 3
17 2.92
18 2.87
19 2.81
20 2.75
21 2.71
22 2.68
23 2.62
24 2.6
25 2.58
26 2.56
27 2.53
28 2.51
29 2.49
30 2.45
31 2.42
32 2.43
33 2.41
34 2.39
35 2.38
36 2.37
37 2.36
38 2.34
39 2.32
40 2.3
41 2.28
42 2.27
43 2.26
44 2.25
45 2.25
46 2.24
47 2.23
48 2.22
49 2.22
50 2.22
};
\addlegendentry{Ours (unbiased)}
\end{axis}

\end{tikzpicture}}}}\hfil 
\subfloat[ImageNet]{\scalebox{0.65}{{
\begin{tikzpicture}

\definecolor{dodgerblue0143213}{RGB}{0,143,213}
\definecolor{goldenrod22917456}{RGB}{229,174,56}
\definecolor{lightgray203}{RGB}{203,203,203}
\definecolor{lightgray204}{RGB}{204,204,204}
\definecolor{tomato2527948}{RGB}{252,79,48}
\definecolor{whitesmoke240}{RGB}{240,240,240}
\definecolor{olivedrab10914479}{RGB}{109,144,79}

\begin{axis}[
axis background/.style={fill=white},
axis line style={whitesmoke240},
legend cell align={left},
legend style={fill opacity=0.8, draw opacity=1, text opacity=1, draw=lightgray204, fill=white, at={(0.97,1.2)}},
tick align=outside,
tick pos=left,
x grid style={lightgray203},
xlabel={Number of samples},
xmajorgrids,
xmin=-1.45, xmax=52.45,
xtick style={color=black},
y grid style={lightgray203},
ylabel={MAPE (\%)},
ymajorgrids,
ymin=0, ymax=32.4045,
ytick style={color=black},
yscale=0.8
]
\addplot [ultra thick, goldenrod22917456, mark size=1.5pt, mark=square*]
table {%
2 23.6
3 23.7
4 23.2
5 22.6
6 21.8
7 21.4
8 20.9
9 20.4
10 20
11 19.4
12 18.9
13 18.4
14 18.1
15 17.7
16 17.4
17 17
18 16.8
19 16.4
20 16.2
21 15.9
22 15.7
23 15.4
24 15.2
25 15
26 14.7
27 14.4
28 14.3
29 14.1
30 13.9
31 13.7
32 13.5
33 13.4
34 13.2
35 13
36 12.9
37 12.8
38 12.7
39 12.4
40 12.4
41 12.2
42 12.1
43 12
44 11.8
45 11.7
46 11.7
47 11.5
48 11.4
49 11.2
50 11.1
};
\addlegendentry{DB Index}

\addplot [ultra thick, tomato2527948, mark size=1.5pt, mark=diamond*]
table {%
2 31.2
3 21.4
4 16.3
5 13.3
6 11.6
7 10.3
8 9.25
9 8.57
10 7.58
11 7.25
12 6.66
13 6.27
14 6.22
15 5.9
16 5.48
17 5.32
18 5.09
19 4.76
20 4.69
21 4.6
22 4.5
23 4.32
24 4.26
25 4.07
26 3.9
27 3.91
28 3.81
29 3.74
30 3.63
31 3.57
32 3.59
33 3.56
34 3.44
35 3.38
36 3.28
37 3.27
38 3.16
39 3.18
40 3.09
41 3.03
42 2.99
43 2.96
44 2.91
45 2.87
46 2.89
47 2.9
48 2.91
49 2.8
50 2.79
};
\addlegendentry{Cross-validation}
\addplot [ultra thick, olivedrab10914479, mark=triangle*, mark size=1.5pt]
table {%
2 24.1
3 24.1
4 23.1
5 21.8
6 20
7 18.5
8 16.9
9 15.5
10 14.3
11 13.2
12 12.2
13 11.3
14 10.6
15 9.95
16 9.35
17 8.86
18 8.38
19 7.87
20 7.37
21 7.02
22 6.65
23 6.29
24 5.98
25 5.71
26 5.48
27 5.19
28 4.95
29 4.75
30 4.58
31 4.39
32 4.24
33 4.08
34 3.99
35 3.87
36 3.75
37 3.63
38 3.49
39 3.38
40 3.26
41 3.18
42 3.09
43 3.03
44 2.95
45 2.89
46 2.82
47 2.78
48 2.76
49 2.72
50 2.64
};
\addlegendentry{Ours (biased)}
\addplot [ultra thick, dodgerblue0143213, mark size=1.5pt, mark=*]
table {%
2 26.4
3 19.1
4 14.1
5 13.8
6 10.7
7 9.15
8 8.51
9 8.07
10 7.31
11 7.07
12 6.8
13 6.5
14 6.37
15 6.07
16 5.81
17 5.55
18 5.37
19 5.32
20 5.21
21 5.02
22 5.03
23 4.95
24 4.86
25 4.78
26 4.78
27 4.66
28 4.62
29 4.69
30 4.61
31 4.53
32 4.47
33 4.44
34 4.39
35 4.34
36 4.35
37 4.3
38 4.25
39 4.26
40 4.23
41 4.29
42 4.22
43 4.24
44 4.2
45 4.13
46 4.17
47 4.22
48 4.17
49 4.14
50 4.14
};
\addlegendentry{Ours (unbiased)}
\end{axis}

\end{tikzpicture}}}} 

\subfloat[VGG-Flower]{\scalebox{0.65}{{
\begin{tikzpicture}

\definecolor{dodgerblue0143213}{RGB}{0,143,213}
\definecolor{goldenrod22917456}{RGB}{229,174,56}
\definecolor{lightgray203}{RGB}{203,203,203}
\definecolor{lightgray204}{RGB}{204,204,204}
\definecolor{tomato2527948}{RGB}{252,79,48}
\definecolor{whitesmoke240}{RGB}{240,240,240}
\definecolor{olivedrab10914479}{RGB}{109,144,79}

\begin{axis}[
axis background/.style={fill=white},
axis line style={whitesmoke240},
legend cell align={left},
legend style={fill opacity=0.8, draw opacity=1, text opacity=1, draw=lightgray204, fill=white, at={(0.97,1.2)}},
tick align=outside,
tick pos=left,
x grid style={lightgray203},
xlabel={Number of samples},
xmajorgrids,
xmin=1.1, xmax=20.9,
xtick style={color=black},
y grid style={lightgray203},
ylabel={MAPE (\%)},
ymajorgrids,
ymin=0, ymax=40.966,
ytick style={color=black},
yscale=0.8
]
\addplot [ultra thick, goldenrod22917456, mark=square*]
table {%
2 50.7
3 48.8
4 46.7
5 44.9
6 43.1
7 41.9
8 40.6
9 39.5
10 38.6
11 37.4
12 36.4
13 35.4
14 34.6
15 33.8
16 33
17 32.3
18 31.6
19 31
20 30.5
};
\addlegendentry{DB Index}

\addplot [ultra thick, tomato2527948, mark=diamond*]
table {%
2 23.5
3 13.7
4 10.7
5 8.85
6 7.44
7 6.77
8 6.38
9 5.83
10 5.55
11 5.13
12 4.92
13 4.64
14 4.51
15 4.42
16 4.4
17 4.21
18 4.02
19 3.98
20 3.98
};
\addlegendentry{Cross-validation}

\addplot [ultra thick, olivedrab10914479, mark=triangle]
table {%
2 63.8
3 49.2
4 41.3
5 36.4
6 32.9
7 30.7
8 28.9
9 27.4
10 26
11 24.9
12 24.1
13 23.2
14 22.6
15 22
16 21.5
17 20.9
18 20.6
19 20.3
20 20
};
\addlegendentry{Ours (biased)}
\addplot [ultra thick, dodgerblue0143213, mark=*]
table {%
2 12.8
3 8.61
4 7.1
5 6.38
6 5.3
7 5.01
8 4.75
9 4.56
10 4.36
11 4.14
12 4.03
13 3.91
14 3.89
15 3.81
16 3.81
17 3.67
18 3.59
19 3.53
20 3.51
};
\addlegendentry{Ours (unbiased)}
\end{axis}

\end{tikzpicture}}}}\hfil   
\subfloat[CUB-200-2011]{\scalebox{0.65}{{
\begin{tikzpicture}

\definecolor{dodgerblue0143213}{RGB}{0,143,213}
\definecolor{goldenrod22917456}{RGB}{229,174,56}
\definecolor{lightgray203}{RGB}{203,203,203}
\definecolor{lightgray204}{RGB}{204,204,204}
\definecolor{tomato2527948}{RGB}{252,79,48}
\definecolor{whitesmoke240}{RGB}{240,240,240}
\definecolor{olivedrab10914479}{RGB}{109,144,79}

\begin{axis}[
axis background/.style={fill=white},
axis line style={whitesmoke240},
legend cell align={left},
legend style={
  fill opacity=0.8,
  draw opacity=1,
  text opacity=1,
  draw=lightgray204,
  fill=white, at={(0.97,1.2)}
},
tick align=outside,
tick pos=left,
x grid style={lightgray203},
xlabel={Number of samples},
xmajorgrids,
xmin=-0.15, xmax=25.15,
xtick style={color=black},
y grid style={lightgray203},
ylabel={MAPE (\%)},
ymajorgrids,
ymin=4.289, ymax=32.691,
ytick style={color=black},
yscale=0.8
]

\addplot [ultra thick, goldenrod22917456, mark size=1.5pt, mark=square*]
table {%
2 17.3
3 15.3
4 12.4
5 10.9
6 9.67
7 9.31
8 8.94
9 8.75
10 8.79
11 9.38
12 10
13 11.1
14 11.8
15 12.5
16 13.9
17 15.1
18 16
19 16.9
20 17.6
21 18.3
22 18.8
23 19.8
24 20.2
};
\addlegendentry{DB Index}
\addplot [ultra thick, tomato2527948, mark size=1.5pt, mark=diamond*]
table {%
2 30.6
3 22.2
4 16
5 14.2
6 12.7
7 11.7
8 10.3
9 9.41
10 8.7
11 8.31
12 8.04
13 7.5
14 7.25
15 6.95
16 6.6
17 6.49
18 6.45
19 6.36
20 6.31
21 6.1
22 6.03
23 5.93
24 5.88
};
\addlegendentry{Cross-validation}

\addplot [ultra thick, olivedrab10914479, mark=triangle*, mark size=1.5pt]
table {%
2 26
3 25.6
4 25
5 24.2
6 23.1
7 21.9
8 20.6
9 19.6
10 18.4
11 17.3
12 16.3
13 15.4
14 14.9
15 14.3
16 13.7
17 13.2
18 12.6
19 12.2
20 11.6
21 11.2
22 10.8
23 10.5
24 10.1
};
\addlegendentry{Ours (biased)}
\addplot [ultra thick, dodgerblue0143213, mark size=1.5pt, mark=*]
table {%
2 25.5
3 18.4
4 15
5 12.8
6 11.9
7 10.7
8 9.82
9 9.2
10 8.45
11 8.26
12 8.03
13 7.58
14 7.44
15 7.12
16 6.95
17 6.74
18 6.65
19 6.55
20 6.57
21 6.27
22 6.14
23 6.11
24 6.02
};
\addlegendentry{Ours (unbiased)}
\end{axis}

\end{tikzpicture}}}}\hfil
\caption{Prediction error of different generalization predictors over $10^3$ few-shot classification problems with 10 classes. Figures (a,b,c) are in-domain datasets and Figures (d,e) are cross-domain datasets. We plot the Mean-Absolute-Percentage Error against the number of samples and compare our method to cross-validation and Davies–Bouldin Index.}
\label{fig:bigResults10ways}
\end{figure}
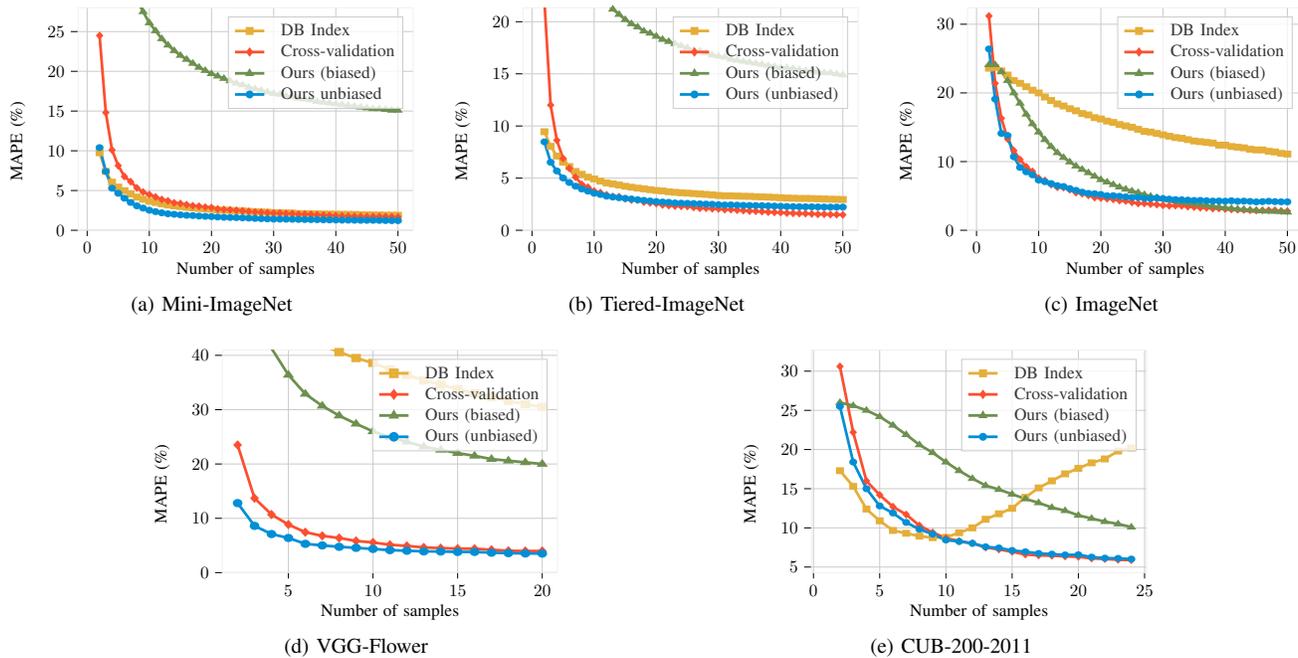

We experiment with different number of classes and show the performance of our method compared to DB-Index and cross-validation.
In Figure~\ref{fig:bigResults10ways} we show the results on 10-class few-shot problems. The trend is similar for different configurations of classes. We did not include DTD in the 10-class problems since the original dataset only has 7 classes. Regarding the performance, our method performs better on average than the alternatives. We only use a matrix with shared isotropic covariance given that the dimensionality of the projected space is 9 for 10-class problems.

\begin{figure}[htbp]
    \centering
    \scalebox{0.8}{\input{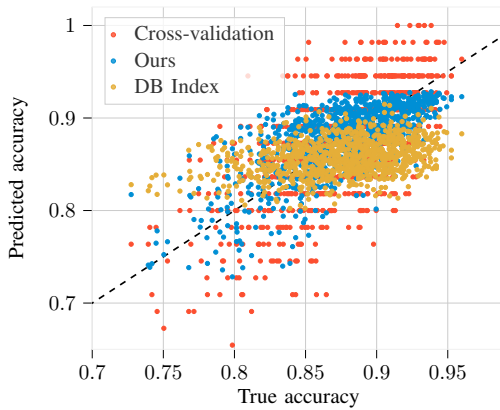}}
    \caption{Scatter plot of 5-class few-shot problems with 10 samples per class from Mini-ImageNet. Each point represents a different problem with a true ground-truth accuracy plotted against the predicted accuracy from the different methods.}
    \label{fig:resultsScatterMiniImagenet}
\end{figure}

In Figure~\ref{fig:resultsScatterMiniImagenet} we include an additional scatter plot for Mini-ImageNet. The predictions from our method are aligned with the ground truth accuracies and are less scattered than the cross-validation approach. 

